\documentclass[journal]{IEEEtran}

\usepackage{amsmath,amsfonts}
\usepackage{algorithmic}
\usepackage{algorithm}
\usepackage{array}
\usepackage[caption=false,font=normalsize,labelfont=sf,textfont=sf]{subfig}
\usepackage{textcomp}
\usepackage{stfloats}
\usepackage{url}
\usepackage{verbatim}
\usepackage{graphicx}
\usepackage{cite}
\usepackage{multirow}
\usepackage{threeparttable}
\usepackage{makecell}
\usepackage[table]{xcolor}
\usepackage{graphicx} 
\usepackage{float}
\usepackage{arydshln} 
\usepackage{hyperref}
\usepackage{bbding} 
 
\begin{document}

\title{Annotation Guidelines-Based Knowledge Augmentation: Towards Enhancing Large Language Models for Educational Text Classification}

\author{
\thanks{This work was financially supported by the National Science and Technology Major Project (2022ZD0117101), National Natural Science Foundation of China (62377016, 62293554, 62293550), Hubei Provincial Natural Science Foundation of China (2023AFA020), and China Scholarship Council (202306770067). \textit{(Corresponding author: Zhi Liu.)} }
Shiqi Liu, Sannyuya Liu,~\IEEEmembership{Member,~IEEE}, Lele Sha, Zijie Zeng, Dragan Ga\v{s}evi\'{c}, and Zhi Liu 

\thanks{Shiqi Liu, Sannyuya Liu, and Zhi Liu are with the National Engineering Research Center of Educational Big Data, Faculty of Artificial Intelligence in Education, Central China Normal University, Wuhan 430079, China (e-mail: liushiqi@mails.ccnu.edu.cn; liusy027@ccnu.edu.cn;  zhiliu@mail.ccnu.edu.cn).
} 
\thanks{Lele Sha, Zijie Zeng, and Dragan Ga\v{s}evi\'{c} are with the Centre for Learning Analytics, Faculty of Information Technology, Monash University, Clayton, VIC 3800, Australia (e-mail: lele.sha@monash.edu; zijie.zeng@monash.edu; dragan.gasevic@monash.edu)}


}

\markboth{IEEE TRANSACTIONS ON LEARNING TECHNOLOGIES,~Vol.~14, No.~8, August~2024}%
{Shell \MakeLowercase{\textit{et al.}}: A Sample Article Using IEEEtran.cls for IEEE Journals}


\maketitle

\begin{abstract} 

Various machine learning approaches have gained significant popularity for the automated classification of educational text to identify indicators of learning engagement -- i.e. learning engagement classification (LEC). LEC can offer comprehensive insights into human learning processes, attracting significant interest from diverse research communities, including Natural Language Processing (NLP), Learning Analytics, and Educational Data Mining. Recently, Large Language Models (LLMs), such as ChatGPT, which are considered promising technologies for Artificial General Intelligence (AGI), have demonstrated remarkable performance in various NLP tasks. However, their comprehensive evaluation and improvement approaches in LEC tasks have not been thoroughly investigated. In this study, we propose the Annotation Guidelines-based Knowledge Augmentation (AGKA) approach to improve LLMs. AGKA employs GPT 4.0 to retrieve label definition knowledge from annotation guidelines, and then applies the random under-sampler to select a few typical examples. Subsequently, we conduct a systematic evaluation benchmark of LEC, which includes six LEC datasets covering behavior classification (question and urgency level), emotion classification (binary and epistemic emotion), and cognition classification (opinion and cognitive presence). The study results demonstrate that AGKA can enhance non-fine-tuned LLMs, particularly GPT 4.0 and Llama 3 70B. GPT 4.0 with AGKA few-shot outperforms full-shot fine-tuned models such as BERT and RoBERTa on simple binary classification datasets. However, GPT 4.0 lags in multi-class tasks that require a deep understanding of complex semantic information. Notably, Llama 3 70B with AGKA is a promising combination based on open-source LLM, because its performance is on par with closed-source GPT 4.0 with AGKA. In addition, LLMs struggle to distinguish between labels with similar names in multi-class classification. Our results provide a valuable benchmark for evaluating LEC models and highlight the effectiveness of AGKA for LLMs in education. Data and code used in our research are available at \url{https://github.com/AnonymousGithubLink}. 

\end{abstract}


\vspace{30pt}

\section{Introduction}

\IEEEPARstart{O}{ver} the past decade, online and blended education has experienced significant global growth. For instance, Udemy, a leading platform for Massive Open Online Courses (MOOCs), has provided over 202,000 online courses to more than 662 million learners as of July 2023 \cite{202KCourses662M2023}. This expansion has led to a surge in learner-generated texts across various communication channels in online learning, including discussion forums, course reviews, social media, and chatbots. Analyzing these unstructured texts enables instructors to understand learners and learning processes, which is critical for providing timely and personalized support. This analysis is beneficial for improving learners' outcomes and retention rates. However, the manual analysis of these texts is labor-intensive and time-consuming, limiting instructors' ability to keep up with the rapid data growth. Consequently, Educational Text Classification (ETC) techniques, which can process large-scale texts automatically in real-time, have become essential and have attracted considerable attention from diverse research communities, including Natural Language Processing (NLP), Learning Analytics, and Educational Data Mining \cite{almatrafiSystematicReviewDiscussion2019a}.

Learning Engagement Classification (LEC), a subfield of ETC, can assist researchers in identifying learners' learning engagement states and implementing customized instructional interventions  \cite{liuAutomatedDetectionEmotional2022,zhengImpactLearningAnalytics2023}. LEC encompasses a range of classification tasks, including aspects of behavior, emotion, and cognition from learner-generated texts \cite{martinOnlineLearnerEngagement2022,liuDualfeatureembeddingsbasedSemisupervisedLearning2022}. For instance, in a MOOC forum, learners may express confusion about the course reading material (epistemic emotion classification) \cite{hanIdentifyingPatternsEpistemic2021} and subsequently decide to seek assistance (questioning behavior classification) \cite{agrawal2015youedu}. After receiving a response from the instructor, learners re-analyze these materials and gain new insights that reflect their higher-order cognition (cognitive presence classification) \cite{gasevicExternallyfacilitatedRegulationScaffolding2015c}. In recent years, LEC studies have explored the application of deep learning-based methods, including Convolutional Neural Network (CNN), Recurrent Neural Network (RNN), and transformer-based pre-trained models such as Bidirectional Encoder Representations from Transformers (BERT) and RoBERTa \cite{liuAutomatedDetectionEmotional2022,shaLatestGreatestComparative2023a}. Nevertheless, these models still require fine-tuning by technologists and underperform when annotated data is insufficient.

Recently, Large Language Models (LLMs), such as ChatGPT, are considered promising technologies for Artificial General Intelligence (AGI) \cite{changSurveyEvaluationLarge}. They achieved remarkable success in various Natural Language Processing (NLP) tasks \cite{qinChatGPTGeneralPurposeNatural2023a,zhangSentimentAnalysisEra2023}. Instruction fine-tuned LLMs, such as GPT 4.0 and Llama 3, can function as general text classifiers by following natural language prompts. They can perform well on novel text classification tasks without requiring fine-tuning on annotated data \cite{chungScalingInstructionFinetunedLanguage2024}. This makes LLMs more accessible for research and practice compared to models that require fine-tuning. Recent studies have investigated the potential of LLMs for LEC, including the classification of classroom dialogue behavior \cite{wangCanChatgptDetect2023b} and social emotions \cite{houPromptbasedFinetunedGPT2024a}.

However, LLMs with vanilla prompts that lack domain knowledge exhibit limitations in numerous text classification tasks \cite{qinChatGPTGeneralPurposeNatural2023a}. For example, the vanilla prompt \textit{Given the [text], assign an emotion label from ["Curiosity", "Confusion", ...]} only considers the name string of a label but overlooks the detailed definition of a label. This raises two major challenges. First, a label name may have different semantics in different tasks and contexts. Second, a label name is insufficient to describe complex or unusual labels. Studies have found that the lack of domain-specific knowledge may lead LLMs to generate error results, which can undermine the trustworthiness of LLMs on LEC tasks \cite{chenBenchmarkingLargeLanguage2024a}. 
To address these limitations, researchers have explored various approaches to augment LLMs with additional knowledge. One such approach is Retrieval-Augmented Generation (RAG) \cite{chenBenchmarkingLargeLanguage2024a}, which involves retrieving relevant information from external knowledge sources and incorporating it into the prompt during the generation process. Another approach is to utilize annotation guidelines \cite{sainzGoLLIEAnnotationGuidelines2023a}, which provide detailed explanations of the labels. By incorporating the information from annotation guidelines into the prompts, LLMs can gain a better understanding of the label semantics and improve their classification performance.

In contrast, during the human annotation process of text classification data, annotators follow the label definitions and typical text-label pairs provided in the annotation guidelines to label unknown data \cite{ideHandbookLinguisticAnnotation2017}. 
Annotation guidelines provide detailed explanations of the labels, along with examples and uncommon cases, to ensure consistent and accurate labeling of the data \cite{sainzGoLLIEAnnotationGuidelines2023a,qiaoReasoningLanguageModel2023}. In the context of LEC, the use of annotation guidelines helps to capture the subtle nuances between labels such as "Curiosity" and "Confusion," which may be challenging to distinguish based on the label names alone. Inspired by this human-labeled approach, in this study, we propose a novel approach, i.e., \textbf{A}nnotation \textbf{G}uidelines-based \textbf{K}nowledge \textbf{A}ugmentation (\textbf{AGKA}), which aims to improve LLMs by retrieving knowledge about label definitions from annotation guidelines and selecting a few typical examples. Furthermore, we perform a comprehensive evaluation of the performance of LLMs on six LEC tasks.

The main contributions and findings of this study are as follows:
\begin{enumerate}

\item We propose the LEC benchmark to comprehensively evaluate LLMs in six LEC  datasets, focusing on the classification of behavior, emotion, and cognition in educational text data. The evaluated models are six non-fine-tuned LLMs (series of GPT, Llama 3, and Mistral) and two fine-tuned models (BERT and RoBERTa). 

\item We propose AGKA, a novel approach to improve LLMs by retrieving knowledge from annotation guidelines and randomly under-sampling typical few-shot examples. Our results show that AGKA increases the weighted F1 scores of LLMs on LEC tasks by up to 8.48\%.

\item GPT 4.0 with AGKA outperformed full-shot fine-tuned models on three binary classification tasks, so AGKA may be adopted to boost LEC classification performance when fine-tuning is not viable. 

\item Llama 3 70B with AGKA performed on par with GPT 4.0, indicating that open-sourced models are catching up to closed-sourced models.

\item Even with AGKA, LLMs still struggle with multi-class classification tasks that require a deep understanding of complex semantic information, such as epistemic emotion and cognitive presence.

\end{enumerate}

\section{Related Works}
\subsection{Learning Engagement Classification} 

The learning engagement implicit in textual products of learning (e.g., discussion posts) can reflect a learner's multidimensional learning state, including behavior, emotion, and cognition \cite{huangExaminingRelationshipPeer2023,martinOnlineLearnerEngagement2022}. LEC seeks to identify these aspects in the textual products of learners' activities. This area has attracted considerable interest and continues to be an important research area for two main reasons. First, understanding the human learning process through textual data is essential in educational research to investigate learning patterns \cite{almatrafiSystematicReviewDiscussion2019a}. For example, LEC for MOOC forums can reveal interactions between emotion and cognition \cite{liuAutomatedDetectionEmotional2022}. Second, the practical applications of LEC are extensive, especially due to the rapid growth of learner-generated content, including sentiment analysis of MOOC reviews, opinion mining in forum discussions, and discourse analysis for chatbots \cite{fengEmotionAnalysisDataset2022,linItGoodMove2022d}.

Traditional LEC methods, based on machine learning techniques such as Support Vector Machine (SVM) and Random Forest (RF), require much time from domain experts to extract hand-crafted features and generally yield suboptimal performance \cite{kovanovicAutomatedContentAnalysis2016b,barbosaAutomaticCrosslanguageClassification2020}. Recently, deep learning-based approaches have been employed to automatically learn feature representations for LEC tasks from text. These methods include CNN, RNN, and BERT \cite{shaLatestGreatestComparative2023a,liuLookingMOOCDiscussion2022,yanPracticalEthicalChallenges2023a,liuAutomatedDetectionEmotional2022}. However, they underperform on datasets with a small number of annotated datasets.
In addition, we need to fine-tune the model specific to each dataset rather than using a universal model that does not require fine-tuning.

\subsection{Large Language Models in LEC}

The development of LLMs has significantly advanced performance on various NLP tasks \cite{laskarSystematicStudyComprehensive2023}, e.g., text classification, which is also one of the fundamental capabilities of LLMs \cite{changSurveyEvaluationLarge}. Notable models include GPT 3.5, GPT 4.0, Llama 3 \cite{touvronLlamaOpenEfficient2023a}, and Mistral \cite{jiangMistral7B2023}, which are pre-trained on large text corpora and various training methods such as instruction tuning and Reinforcement Learning from Human Feedback (RLHF) \cite{ouyangTrainingLanguageModels2022a}. Studies show that LLMs exhibit remarkable performance in zero-shot and few-shot learning\cite{sahoo2024systematic}. Zero-shot learning refers to the model being instructed to perform a task without any additional training or examples. In contrast, few-shot learning involves a model being provided with a small number of examples before being instructed to perform the task\cite{sahoo2024systematic}. LLMs have shifted the NLP paradigm from fine-tuning specific models to using prompts. Prompts are task-specific instructions that guide the model to generate the desired output \cite{qinChatGPTGeneralPurposeNatural2023a}.

Several studies have applied LLMs to address LEC tasks in education, such as classifying dialogue behaviors and social emotions \cite{houPromptbasedFinetunedGPT2024a}. Comparative evaluations show that while GPT 3.5 may not be able to match the peformance of full-shot fine-tuned BERT, it shows potential for identifying specific social interactions \cite{wangCanChatGPTDetect2023a}. 
However, studies have found that lack of domain-specific knowledge may lead LLMs to generate errors and illusions, which can limit their performance in LEC tasks \cite{chenBenchmarkingLargeLanguage2024a,sainzGoLLIEAnnotationGuidelines2023a}. 
Previous studies have only evaluated a limited number of private datasets and closed-source LLMs, which may not provide a comprehensive understanding of the potential of LLMs in LEC. Therefore, one of the goals of this study is to conduct a comprehensive evaluation of open-source LLMs and LEC datasets.

\begin{figure*}
    \centering
    \includegraphics[width=0.6\linewidth]{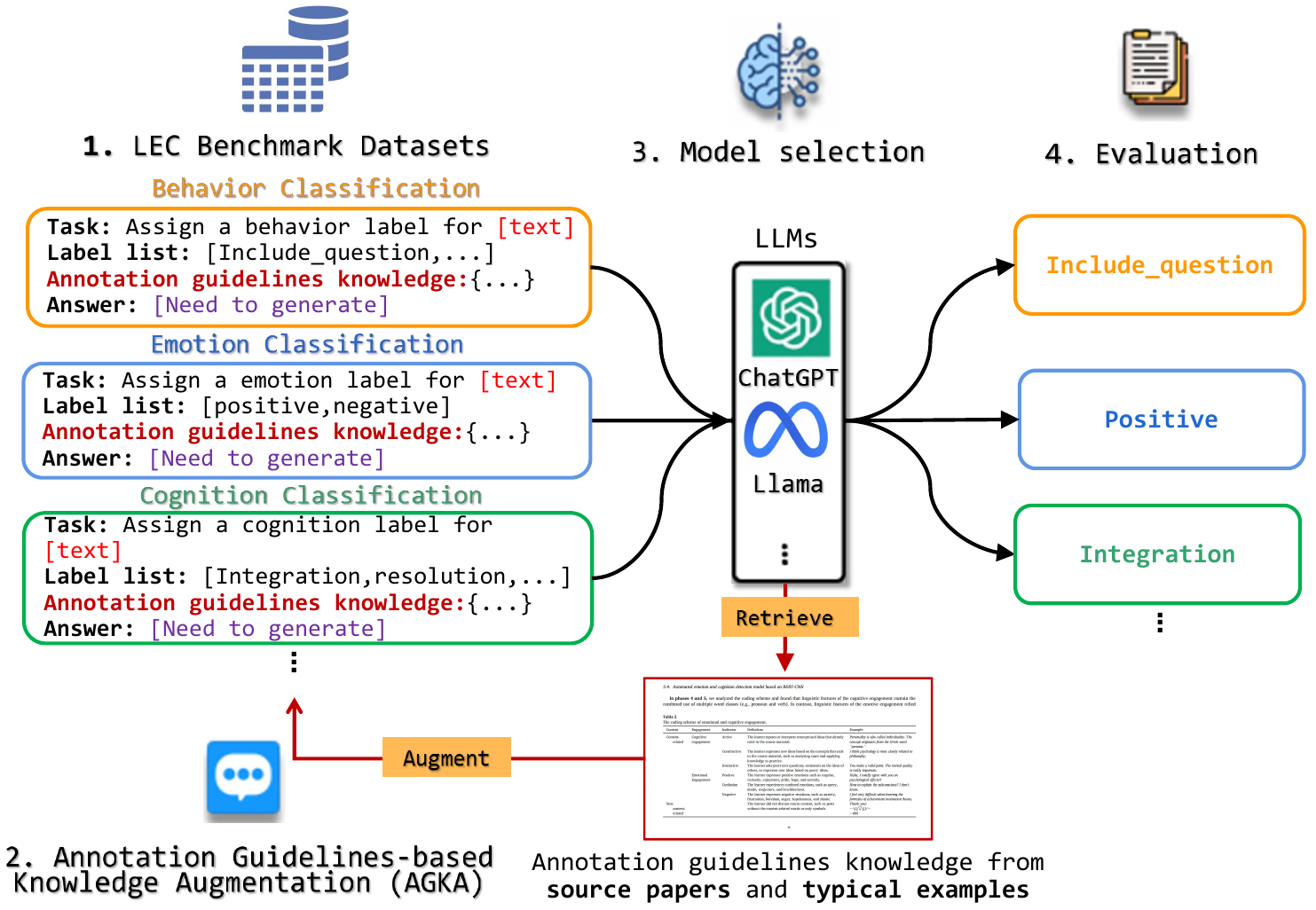}
    \caption{The universal process of LEC based on LLMs with AGKA.}
    \label{figure_method}
\end{figure*}

\subsection{Prompt Engineering}
Prompt engineering has emerged as an effective method of adapting LLMs to specific tasks without the need for fine-tuning \cite{changSurveyEvaluationLarge}. By designing appropriate prompts, LLMs can be directed to perform a variety of tasks such as text classification, generation, and reasoning \cite{laskarSystematicStudyComprehensive2023}. For example, ChatGPT is shown to outperform crowd workers in several annotation tasks, including stance, topic, and frame detection \cite{gilardiChatGPTOutperformsCrowd2023a}. In addition, prompt engineering has proven successful in a wide range of general text classification tasks, such as news classification and sentiment analysis \cite{sunTextClassificationLarge2023,gretzZeroshotTopicalText2023a}. 

Recent studies also demonstrated that augmenting LLMs with domain knowledge can significantly improve performance on domain-specific tasks through techniques such as RAG \cite{lewisRetrievalAugmentedGenerationKnowledgeIntensive2020a}, annotation guidelines knowledge summarization \cite{sainzGoLLIEAnnotationGuidelines2023a}, and  common knowledge reasoning \cite{liu-etal-2022-generated}. Therefore, prompt engineering is a promising strategy to enhance the capabilities of LLMs in LEC.

\section{Method}

The methodology of this study consists of four phases, as shown in \textbf{Fig \ref{figure_method}}.
strategies. 

\subsection{LEC Benchmark Datasets}

As shown in \textbf{Table \ref{tab:table1}}, this study was conducted to provide a comprehensive review of a wide range of LEC tasks and grouped them into three types of learning engagement aspects including behavior, emotion, and cognition \cite{xuEffectsTeacherRole2020,martinOnlineLearnerEngagement2022}. We collected six datasets for these LEC tasks, with two datasets per aspect. These datasets have been widely used and are highly representative to comprehensively evaluate the LEC ability of LLMs. The datasets are selected from previously published studies, are all in English language, and contain transcripts of discussions on online forums. They cover a range of disciplines, including computer science, management, mathematics, economics, and medicine \cite{agrawal2015youedu,demszkyGoEmotionsDatasetFineGrained2020,gasevicExternallyfacilitatedRegulationScaffolding2015c}.

\begin{table*}
\centering
\caption{Description for evaluation Datasets.}
\label{tab:table1}
\resizebox{!}{3.5CM}{
\begin{threeparttable}         
\begin{tabular}{>{\raggedright\arraybackslash}p{0.08\linewidth}>{\raggedright\arraybackslash}p{0.08\linewidth}l >{\raggedright\arraybackslash}p{0.2\linewidth}l l l l >{\raggedright\arraybackslash}p{0.2\linewidth}l}    
\hline \textbf{Task} & \textbf{Dataset} & \textbf{Language} & \textbf{Describe} &\textbf{\#Train} \tnote{1}&\textbf{\#Val}&\textbf{\#Test} & \textbf{Class} &\textbf{Label (\%)}  \tnote{2} & \textbf{Cite}\\ \hline
\rowcolor{gray!20} Behavior Classification & Urgency Level  & English & Detect whether the text needs to be handled immediately by the instructor.&20724&4440&4440 & 2  & \texttt{High\_urgency (19.60\%),Low\_urgency (80.40\%), }&\cite{agrawal2015youedu}\\
& Question Detection  & English & Detect whether the text contains a question.&20724&4440&4440 & 2 & \texttt{Include\_question (20.50\%), No\_question(79.50\%)} & \cite{agrawal2015youedu}\\
\rowcolor{gray!20} Emotion Classification & Binary Emotion & English & Detect positive or negative emotions in student discussions of the text.  &11413&2445&2445 & 2 &\texttt{Positive (70.90\%), Negative (29.10\%)} & \cite{agrawal2015youedu} \\ 
 & Epistemic Emotion & English & Learners experience complex fine-grained emotions during the knowledge construction process. &1581&338&338& 6  & \texttt{Neutral(73.90\%), Surprise(4.40\%), Curiosity(7.80\%), Enjoyment(4.20\%)}  \texttt{Anxiety(2.80\%), Confusion(4.90\%)} &\cite{demszkyGoEmotionsDatasetFineGrained2020}\\
\rowcolor{gray!20} Cognitive Classification & Opinion Detection & English & Determine if the text contains a learner's subjective opinion. &1748 &20724&4440 & 2  & \texttt{Contain\_opinion (45.10\%), No\_opinion (54.90\%)} & \cite{agrawal2015youedu}\\
 & Cognitive Presence & English & Identify the cognitive classes of a discussion text. &1223&262&262 & 5  & \texttt{Triggering\_Event (17.60\%), Exploration(39.20\%) } \texttt{Integration(29.00\%), Resolution(6.10\%), Other(8.10\%)}& \cite{gasevicExternallyfacilitatedRegulationScaffolding2015c}\\
 \hline
\end{tabular}
         \begin{tablenotes}  
        \footnotesize   
        \item[1] The training, validation, and test datasets are in the ratio of 70\%:15\%:15\% for full-shot fine-tuned models, including BERT and RoBERTa. 
        \item[2] The ratio of each label in the entire dataset.   
      \end{tablenotes}  
    \end{threeparttable} 
}
\end{table*}

\subsubsection*{\bf Datasets in Behavior Classification}
Dialogue acts or behaviors are interpreted as the basic units of a conversation, more fine-grained than utterances and characterized by specific communicative functions\cite{blacheTwolevelClassificationDialogue2020}. Therefore, in this study, the behavior classification aimed to classify the underlying intention in a given text, determining whether it involved asking a question or reporting a state of urgency \cite{agrawal2015youedu}. Analyzing discussion behavior is a powerful tool for understanding learning patterns in forum discussions or conversations.

\begin{itemize}

\item{\bf Urgency Level.} The urgency level of a discussion post expresses the learner's help-seeking behavior or intent. This study selected a dataset from \cite{agrawal2015youedu}, in which a high urgency forum post means that a learner needed immediate intervention from instructors. In pre-processing the data, this study labeled posts with urgency level 4 or higher as \texttt{High\_urgency (19.30\%)} and others as \texttt{Low\_urgency (80.40\%)}, following the previous studies \cite{almatrafiNeedleHaystackIdentifying2018a,sha2022leveraging}. 

\item{\bf Question Detection.} Question asking is an important learning action.  The dataset about question asking was annotated in the study presented in \cite{agrawal2015youedu}. Its labels include two categories for each online discussion post -- \texttt{Include\_question(20.50\%)} and \texttt{No\_question(79.50\%)}. 

\end{itemize}

\subsubsection*{\bf Datasets of Emotion Classification} Emotion is a key factor in learning, both as a cause (e.g., curiosity) and as an consequence (e.g., enjoyment) of the learning process \cite{hoyEducationalPsychology2016,liuAutomatedDetectionEmotional2022}. Emotion classification aims to classify a given text into predefined emotion categories, such as  binary categories (e.g., positive or negative) or some more complex  emotion categories, such as epistemic emotions (e.g., confused or enjoyment) \cite{liuSentimentAnalysisMining2020,hanIdentifyingPatternsEpistemic2021}. The  datasets used in the current study extend the emotion classification and focus on identifying and understanding a more broad range of human emotional states. The datasets are described as follows:

\begin{itemize}
\item{\bf Binary Emotion.} This dataset is collected from a series of MOOC forum discussions and annotated the study presented in \cite{agrawal2015youedu}. Its text labels include two emotion categories -- \texttt{Positive (70.10\%)} and \texttt{Negative (29.10\%)}. In data pre-processing, we marked posts with sentiment scores less than 4 as \texttt{Negative} and greater than 4 as \texttt{Positive}, which follow \cite{agrawal2015youedu} \cite{leeFewshotEnoughExploring2023}.

\item{\bf Epistemic Emotion.} Epistemic emotions trigger or constrain the knowledge construction process \cite{chevrierExploringAntecedentsConsequences2019,hanIdentifyingPatternsEpistemic2021}. For example, curiosity contributes significantly to higher-order cognitive skills and learning performance \cite{chevrierExploringAntecedentsConsequences2019}. This dataset is sampled and labeled from a discussion forum \cite{demszkyGoEmotionsDatasetFineGrained2020}. We selected texts and labels related to epistemic emotions in this fine-grained emotions dataset, where each text had only one emotion label. Due to the absence of \texttt{Frustration} and \texttt{Boredom} in the given dataset, the current study included six emotion categories -- \texttt{Neutral (73.90\%)}, \texttt{Surprise (4.40\%)}, \texttt{Curiosity (7.80\%)}, \texttt{Enjoyment (4.20\%),} \texttt{Anxiety (2.80\%)}, and \texttt{Confusion (4.90\%)}. 

\end{itemize}

\subsubsection*{\bf Datasets of Cognition Classification} Cognitive level in educational conversation reflects the level of knowledge construction or critical thinking ability shown in learning processes \cite{chiICAPFrameworkLinking2014,liuAutomatedDetectionEmotional2022}. Cognition classification aims to classify a given text into predefined cognitive categories\cite{liuDualfeatureembeddingsbasedSemisupervisedLearning2022}. The datasets used in the current study included the following:
\begin{itemize}

\item{\bf Opinion Detection.} A learner's expression of opinions in a post, rather than simply quoting textbook material, is a signal of the cognitive process \cite{atapattuDetectingCognitiveEngagement2019}. This dataset was collected from a MOOC forum discussion and annotated in the study presened in  \cite{agrawal2015youedu}. Its text labels include binary categories:  \texttt{Contain\_opinion (45.10\%)} --  meaning the post contained subjective opinions, and \texttt{No\_opinion (54.90\%)} -- meaning the post contained only textbook material without subjective opinions. 

\item{\bf Cognitive Presence.}  The dataset is based on the the community of inquiry framework \cite{garrisonCriticalThinkingCognitive2001b} used to measure the quality of critical thinking, high-order thinking processes, and practical inquiry \cite{netoAutomaticContentAnalysis2021}. This dataset comes from a discussion in a MOOC course and has been annotated by \cite{gasevicExternallyfacilitatedRegulationScaffolding2015c}. Cognitive presence consists of four phases: 1) \texttt{Triggering\_Event (17.60\%)} -- the beginning of collective discussions and reflects the initial phase of the inquiry process, such as the presentation of a problem or dilemma; 2) \texttt{Exploration(39.20\%)} -- learners explore possible solutions to a particular problem, such as searching for information and discussing different ideas; 3) \texttt{Integration(29.00\%)} -- learners synthesize new ideas and knowledge through social construction; and 4) \texttt{Resolution(6.10\%)} -- learners solve the original problem, such as evaluating the acquired knowledge or applying it to practical problems. 

\end{itemize}

\begin{figure*}
    \centering
    \includegraphics[width=0.65\linewidth]{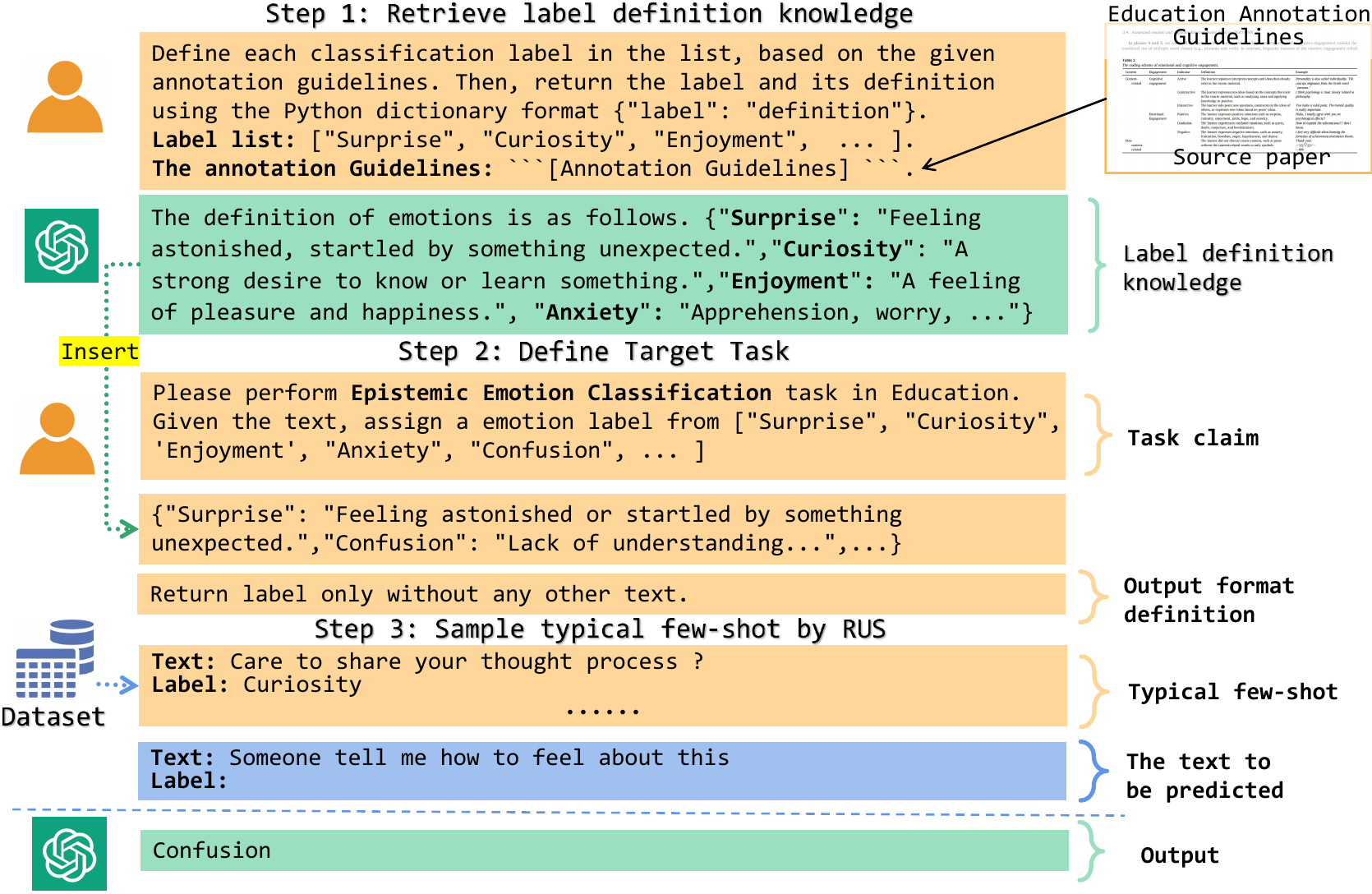}
    \caption{Components of AGKA. \textbf{AGKA} = Task claim + label definition knowledge + Output format definition + (Typical few-shot) + The text to be predicted. 
    \textbf{Vanilla} = Task claim + Output format definition + The text to be predicted. RUS refers to Random Under Sampler.}
    \label{fig_prompt}
\end{figure*}

\subsection{Annotation Guidelines-based Knowledge Augmentation (AGKA) }

\subsubsection{Problem Definition}
Prompt engineering involves the strategic design of task-specific instructions, known as prompts, to instruct model generation without the need for fine-tuning of parameters \cite{sahoo2024systematic}. For a text classification task, given a prompt containing a task instruction $\mathcal{T}$, a test task $\mathcal{X}$ must be solved using a parameterized probabilistic model of an LLM $p_{\rm LLM}$. Our goal is to maximize the likelihood of a target text response $\mathcal{A}$ to solve the test problem, as shown:
\begin{align}
    p(\mathcal{A} \mid \mathcal{X},\mathcal{T})=\prod_{i=1}^{|\mathcal{A}|} p_{\rm LLM} \left ( a_i \mid \mathcal{X},\mathcal{T},a_{< i} \right)
    \label{eq:1}
\end{align}

where $a_i$ and $|\mathcal{A}|$ denote the $i$-th token and the length of the final answer, respectively. For example, when LLMs perform an emotion classification task, the task prompt $\mathcal{T}$ is \textit{Given the text, assign an emotion label from ['Positive', 'Negative']. Return only the label without any other text.}  After the model reads the instruction $\mathcal{T}$ and the input $\mathcal{X}$ \textit{This is such a great way to explain this!}, it is expected to generate the response $\mathcal{A}$ \textit{Positive}.

\subsubsection{Components of AGKA} 
 
As is shown in Fig. \ref{fig_prompt}, AGKA consists of three main steps for each task. In addition, we provide prompt examples of AGKA
in the  Appendix  \ref{APPENDIX}.

\textbf{Step 1: Retrieve Label Definition Knowledge.} We use GPT 4.0 to retrieve the definition of labels from the annotation guide that is collected from the source research article of each dataset. This process involves defining labels in the label list and returning the formatted Python dictionary format \texttt{\{label\_name: definition\}}.

\textbf{Step 2: Define Target Task.} It includes task claim, label definition knowledge, output format definition, and the input text to be predicted. Specifically, the task claim describes the task name and label list; the label definition knowledge is retrieved from step 1; the output format definition controls generation target; and the text to be predicted is selected from the test dataset. For example, the prompt instructs the model to perform the epistemic emotion classification task by assigning an emotion label to a given text from the predefined label list. The model is expected to return only the predicted label without any additional text. 

\textbf{Step 3: Sample Typical Few-Shot by Random Under Sampler (RUS).}  Few-shot provides models with typical examples of \texttt{\textless text, label\textgreater} pairs to improve understanding of a given task, unlike zero-shot prompt where no examples are provided \cite{brown2020language}. To address class imbalance and duplicate sampling, the current study used RUS to under-sample typical shots \cite{chawla2002smote}. By ensuring that the model learns on a more balanced set of examples, RUS helps reduce the bias toward the majority class, leading to better generalization to the minority class. In addition, if the number of samples $S_n$ is less than the number of categories $C_n$, we extract the top $S_n$ from the sample results with $C_n$. The few-shot setting also requires more tokens to contain the examples, which may limit longer input text. To test the effect of sample size on performance, we used RUS to sample a few examples \texttt{[1,5,10]} for each task.

\subsubsection{Prompt Baseline} The vanilla prompt is a baseline for comparison of prompt engineering, which removes the label definition knowledge and few-shot of AGKA. So it contains only the task claim and the output format definition. It follows the definition of vanilla prompt in previous studies \cite{qinChatGPTGeneralPurposeNatural2023a,laskarSystematicStudyComprehensive2023}.

\subsection{Model Selection} 

To evaluate the performance of LLMs, we performed direct inference LLMs on the downstream LEC tasks via prompt engineering without fine-tuning. We selected the state of the art (SOTA) series of models under open and closed-source licenses\footnote{https://chat.lmsys.org/ (as of 1st May 2024)}. The selection was based on an LLM benchmark ranking derived from human evaluation and a general LLM evaluation \cite{zheng2023judging}. 
In addition, we also selected the high-performing fine-tuned models from previous LEC studies \cite{liuMOOCBERTAutomaticallyIdentifying2023,liuAutomatedDetectionEmotional2022}.

\subsubsection*{\bf Non Fine-Tuned Closed-Source LLMs} We chose two LLMs from OpenAI, including GPT 3.5-turbo and the GPT 4.0-turbo\footnote{The experiments used the 25 Jan 2024 version of GPT 3.5-turbo and the GPT 4.0-turbo. It should be noted that future updates could potentially influence the results presented in this paper.}.

\begin{itemize}
\item{\textbf{GPT 3.5-turbo}}. This model is also called ChatGPT and is developed by OpenAI, which can be accessed through the Application Programming Interface (API). 
\item{\textbf{GPT 4.0-turbo}}. GPT 4.0 is the latest version of the GPT available. It uses more recent training data and outperforms GPT 3.5 and other open-source LLMs on both human assessment and general assessment metrics \cite{metaIntroducingMetaLlama}.
\end{itemize}

\subsubsection*{\bf Non-Fine-Tuned Open-Source LLMs} Open-source LLMs can be deployed on private servers, with lower prices and better privacy protections for educational practice and research. We adopted four models from the series of Mistral and Llama 3 because they are open-source and have shown strong zero-shot and few-shot performance, namely Mistral 7 Billion (B), Mixtral 8x7B, Llama 3 8B, and Llama 3 70B. We used APIs to call the models deployed in a private cloud server\footnote{https://www.fireworks.ai/models} because deploying open-source LLMs in the local environment requires heavy hardware.

\begin{itemize}
\item{\textbf{Mistral 7B}}. It is an LLM with 7B parameters developed by Mistral AI \cite{jiang2024mixtral}. We chose its instruction fine-tuned vision, which significantly outperformed 7B and 13B of Llama 1 and 2 \cite{jiang2024mixtral,zheng2023judging}.  
\item{\textbf{Mixtral 8x7B}}. This is a Sparse Mixture of Experts (SMoE) language model with 56B parameters and an improved version based on Mistral 7B. In this model, a router network selects two experts for each token at each layer to process the current state and combine their results. The model fine-tuned to follow instructions, i.e. Mixtral 8x7B-Instruct, outperformed Gemini Pro and Llama 2 70B on all benchmarks evaluated \cite{jiang2024mixtral}.  
\item{\textbf{Llama 3 8B and 70B}} . They represent a new SOTA in a wide range of industry benchmarks for LLMs, compared to the Mistral series and the Llama 2 series \cite{metaIntroducingMetaLlama}. We selected their instruction fine-tuned version.

\end{itemize}

\subsubsection*{\bf Fine-Tuned Models} To compare the performance of LLMs and previous fine-tuned models, we chose BERT and its improved version RoBERTa with few-shot and full-shot fine-tuning as a baseline. In particular, we use only text and label pairs for training them.

\begin{itemize}

\item{\bf BERT}. We choose the BERT base with 110 million parameters \cite{devlinBERTPretrainingDeep2019}, which is widely used and has shown excellent performance on LEC tasks \cite{liuAutomatedDetectionEmotional2022, liuMOOCBERTAutomaticallyIdentifying2023}. 
\item{\bf RoBERTa}.
This is a robustly optimized BERT with 125 million parameters. RoBERTa has been shown to outperform BERT in several NLP benchmarks due to its optimized training regimen that includes longer sequences and more extensive data during pre-training \cite{liu2019roberta}. Many text classification studies regard it as a powerful baseline because of its excellent performance on various text classification tasks\cite{karlTransformersAreShortText2023,sunTextClassificationLarge2023}.

\end{itemize}

\textbf{Random:} It is also included as a baseline for better comparisons, i.e. it assigns a random label to a sample from the label list as a prediction result.

\subsection{Evaluation}

\textbf{Evaluation Metrics.} In evaluating the performance of models, two essential metrics often considered are Accuracy and Weighted F1 Score \cite{zhaoChatAgriExploringPotentials2023a}. All metrics are reported in percentage terms rather than in decimal format.
Accuracy is a fundamental metric, defined as the ratio of correctly predicted observations to total observations. It provides a simple and intuitive measure of overall model performance, but may not always reflect true effectiveness, especially in unbalanced datasets. 



To address this limitation, the weighted F1 is employed, which balances the precision and recall of the classification model, taking into account the relative contribution of each class. It is the weighted harmonic mean of precision and recall, where the weights correspond to the number of true instances for each class, thereby adjusting for class imbalances. 
In the following, the weighted F1 score will be abbreviated to F1 and used as the primary comparison metric.

\begin{table*}[t]
\centering
\caption{Classification Results of Behavior, Emotion, and Cognition}
\label{result_table}
\resizebox{!}{6.8CM}{
\begin{threeparttable}
\begin{tabular}{l |l| l l |l l |l l| l l| l l| l l} \hline 
\multirow{3}{*}{\textbf{Model}} & \multirow{3}{*}{\textbf{Shot}} & \multicolumn{4}{l|}{\textbf{Behavior Classification}} & \multicolumn{4}{l|}{\textbf{Emotion Classification}} & \multicolumn{4}{l}{\textbf{Cognition Classification}} \\ \cline{3-14}
&   & \multicolumn{2}{l|}{\textbf{Urgency}} & \multicolumn{2}{l|}{\textbf{Question}} & \multicolumn{2}{l|}{\textbf{Binary}} & \multicolumn{2}{l|}{\textbf{Epistemic}} & \multicolumn{2}{l|}{\textbf{Opinion}} & \multicolumn{2}{l}{\textbf{Cognitive Presence}} \\ \cline{3-14}
 &  & \textbf{Acc} & \textbf{F1} & \textbf{Acc} & \textbf{F1} & \textbf{Acc} & \textbf{F1} & \textbf{Acc} & \textbf{F1} & \textbf{Acc} & \textbf{F1} & \textbf{Acc} & \textbf{F1} \\ \hline 
\multicolumn{14}{c}{\textbf{Baseline}} \\ \hline 
\rowcolor{gray!20} Random & - & 50.80 & 57.40 & 57.40 & 60.91 & 46.80 & 49.82 & 13.40 & 18.47 & 47.40 & 47.51 & 21.40 & 23.71 \\ 

\hline 
\multicolumn{14}{c}{\textbf{ Fine-Tuned Models}} \\ \hline 
\rowcolor{gray!20} BERT & 1000& 90.47& 82.91& 95.27& 93.40& 93.07& 91.09& 75.47& 43.85& 77.40& 77.65& 54.20& 47.87\\
\rowcolor{gray!20}   & Full& 94.37& 89.75& 95.43& 93.72& 94.53& 92.32& 77.07& \underline{54.80}& 81.40& \textbf{81.30}& 51.87& 47.22\\
RoBERTa & 1000& 91.93& 84.82& 94.93& 92.82& 93.87& 92.26& 74.60& 44.02& 79.67& 79.45& 56.27& \underline{51.19}\\
 & Full& 94.47& 90.31& 95.53& 93.95& 95.80& 94.64& 77.20& \textbf{55.56}& 81.00& \underline{80.89}& 54.67& \textbf{52.51}\\
\hline  
\multicolumn{14}{c}{\textbf{Non-Fine-Tuned LLMs with Vanilla\cite{qinChatGPTGeneralPurposeNatural2023a}}} \\ \hline 
\rowcolor{gray!20} Mistral 7B & 0 & 76.20 & 78.79 & 78.80 & 80.52 & 91.60 & 92.00 & 35.60 & 42.93 & 59.60 & 56.88 & 20.20 & 16.87 \\
Mixtral 8x7B & 0 & 67.40 & 71.71 & 54.40 & 66.48 & 71.80 & 80.25 & 12.40 & 19.27 & 38.60 & 44.10 & 39.60 & 28.34 \\
\rowcolor{gray!20} Llama 3 8B & 0 & 78.20 & 80.03 & 61.80 & 64.35 & 94.20 & 94.26 & 16.80 & 18.07 & 59.00 & 54.87 & 42.40 & 28.38 \\
Llama 3 70B & 0 & 84.20 & 82.38 & 91.40 & 91.85 & 93.20 & 93.90 & 23.40 & 23.58 & 67.20 & 66.83 & 39.80 & 37.84 \\
 \rowcolor{gray!20}  GPT 3.5 & 0 & 84.20 & 84.75 & 88.00 & 88.70 & 92.60 & 92.88 & 29.20 & 32.48 & 48.80 & 36.97 & 38.80 & 33.40 \\
GPT 4.0 & 0 & 83.60 & 78.18 & 94.20 & \underline{94.32}& 93.40 & 93.23 & 19.60 & 19.83 & 63.40 & 62.08 & 40.80 & 36.47 \\
\hline
\multicolumn{14}{c}{\textbf{Non-Fine-Tuned LLMs with AGKA (ours)}} \\ \hline 
\rowcolor{gray!20} Mistral 7B & 0 & 82.60 & 84.55 & 82.60 & 83.41 & 92.00 & 92.25 & 32.40 & 39.89 & 65.80 & 65.72 & 15.00 & 20.19 \\
\rowcolor{gray!20} & 1 & 55.00 & 60.21 & 37.80 & 36.99 & 90.20 & 90.78 & 15.20 & 19.55 & 58.20 & 58.55 & 40.00 & 39.01 \\
\rowcolor{gray!20}  & 5 & 80.20 & 82.16 & 18.00 & 12.94 & 90.20 & 90.83 & 4.20 & 3.23 & 47.20 & 50.80 & 41.80 & 26.74 \\
\rowcolor{gray!20}  & 10 & 67.40 & 72.93 & 16.80 & 18.97 & 90.80 & 92.66 & 3.40 & 2.85 & 10.80 & 17.05 & 37.20 & 24.76 \\
Mixtral 8x7B & 0 & 21.60 & 33.38 & 51.40 & 66.28 & 55.60 & 69.71 & 34.60 & 45.43 & 47.20 & 52.39 & 36.60 & 28.23 \\
 & 1 & 2.00 & 3.84 & 0.00 & 0.00 & 65.80 & 77.86 & 17.00 & 24.24 & 21.60 & 32.32 & 25.20 & 26.13 \\
 & 5 & 68.60 & 76.50 & 0.00 & 0.00 & 50.80 & 64.99 & 17.40 & 24.54 & 31.60 & 39.21 & 22.60 & 21.48 \\
 & 10 & 69.40 & 76.98 & 0.60 & 1.19 & 56.00 & 70.22 & 24.80 & 32.40 & 1.80 & 3.51 & 1.91 & 2.61 \\
 \rowcolor{gray!20}Llama 3 8B & 0& 85.00 & 81.58 & 88.20 & 88.84 & 94.20 & 94.25 & 19.80 & 22.74 & 61.60 & 58.79 & 46.00 & 34.34 \\
 \rowcolor{gray!20}& 1& 60.20 & 65.33 & 73.80 & 77.03 & 90.80 & 91.62 & 12.20 & 9.61 & 52.40 & 43.21 & 42.80 & 36.70 \\
\rowcolor{gray!20} & 5& 32.60 & 46.71 & 15.60 & 16.71 & 2.00 & 3.87 & 20.40 & 27.45 & 45.20 & 28.22 & 22.60 & 19.28 \\
\rowcolor{gray!20} &10 & 1.40 & 2.73 & 1.00 & 1.87 & 0.00 & 0.00 & 18.00 & 22.84 & 1.80 & 3.39 & 42.60 & 26.73 \\
Llama 3 70B & 0& 85.00 & 83.54 & 91.00 & 91.52 & 94.80 & 94.88 & 28.80 & 31.33 & 68.60 & 68.44 & 43.60 & 40.74 \\
 & 1& 86.40 & 86.90 & 89.60 & 90.17 & 93.00 & 93.61 & 19.40 & 16.64 & 71.40 & 71.37 & 32.80 & 31.91 \\
 &5 & 88.60 & 89.25 & 87.00 & 87.87 & 91.80 & 91.89 & 28.20 & 32.61 & 56.20 & 51.26 & 42.80 & 40.84 \\
 & 10& 89.40 & 89.30 & 91.20 & 92.81 & 93.40 & 93.27 & 31.40 & 36.54 & 60.20 & 61.45 & 39.40 & 40.34\\
 \rowcolor{gray!20}  GPT 3.5 & 0 & 89.00 & 88.65 & 92.60 & 93.04 & 92.00 & 91.88 & 37.00 & 43.00 & 50.00 & 39.19 & 42.20 & 34.23 \\
\rowcolor{gray!20} & 1 & 86.60 & 87.32 & 89.40 & 89.98 & 93.00 & 92.94 & 21.80 & 21.36 & 60.40 & 57.11 & 32.60 & 29.76 \\
\rowcolor{gray!20}  & 5 & 90.00& \underline{90.45}& 91.80 & 92.14 & 94.00 & 93.87 & 40.60& 46.98& 57.20 & 51.76 & 30.80 & 25.57 \\
\rowcolor{gray!20}  & 10 & 89.60 & 89.49 & 91.60 & 91.63 & 94.60 & 94.49 & 43.20& 49.10& 65.20 & 64.13 & 23.80 & 17.97 \\
GPT 4.0 & 0 & 85.80 & 82.37 & 94.20& 94.26& 94.60 & 94.51 & 30.00 & 34.97 & 70.00& 69.79& 46.40& 43.18\\ 
 & 1 & 87.80 & 86.91 & 94.80& \textbf{94.88}& 95.60& \underline{95.58}& 19.60 & 19.41 & 72.40& 72.38& 37.20 & 35.57 \\
 & 5 & 89.80& 90.02& 93.00 & 93.05 & 94.40 & 94.31 & 32.80 & 38.34 & 62.80 & 60.82 & 41.60 & 38.68 \\
 & 10 & 90.40& \textbf{90.50}& 93.80 & 93.98 & 95.60& \textbf{95.65}& 27.00 & 32.29 & 66.80 & 66.54 & 44.40& 41.52\\ 
 \hline 
\end{tabular}
 \begin{tablenotes}
        \footnotesize
        \item[1] For the F1 score of each dataset, the best performance is \textbf{bolded} and the second performance is \underline{underlined}, examined by the t-test.
      \end{tablenotes}
  \end{threeparttable}
}
\end{table*}

\textbf{Evaluation Process.} For non-fine-tuned LLMs, we randomly sampled the dataset entailed five times, each time extracting its 15\% as a test data set. Notice that the random seed list for the extraction is the same for all models, so the evaluation data extracted for each model is consistent. 
Moreover, to ensure high confidence and deterministic prediction of the LLMs, their configured \texttt{temperature}, \texttt{max\_tokens}, \texttt{top\_P}, \texttt{presence\_penalty} and \texttt{frequency\_penalty} were set to \texttt{0, 100, 1, 0, and 0,} respectively.

For fine-tuned methods, we randomly extracted five times from the entail dataset for the training, validation, and test datasets in the ratio of 70\%:15\%:15\% (statistics detailed in \textbf{Table \ref{tab:table1}}). In the few-shot fine-tuning setting, we randomly extract the training, validation, and test datasets as in the full-shot fine-tuning setting, but the difference was that we also RUS sample the few-shot number \texttt{[10, 50, 100, 200, 500, 800, 1000]} of samples from the training dataset as a new one. 
For parameter configuration of fine-tuned methods, we employed the Adam optimiser with a learning rate of 1e-5 for BERT and 2e-5 for RoBERTa and a static batch size of 32 for all tasks. In training epochs, we choose 30 times for the full training setting and 50 times for the few-shot training setting \cite{zhangSentimentAnalysisEra2023}. We performed five times with different random seeds for them and report the averages for more robust comparisons. We saved the best model based on the F1 score in the validation set at the end of the epoch. The best performing model in the validation set was used to evaluate the test dataset and save the evaluation metrics. T-test is used to check the significance of differences in model performance measures.

\section{Results}

\subsection{Overall Results Analysis}

\begin{figure*}[t]
    \centering
    \includegraphics[width=0.9\linewidth]{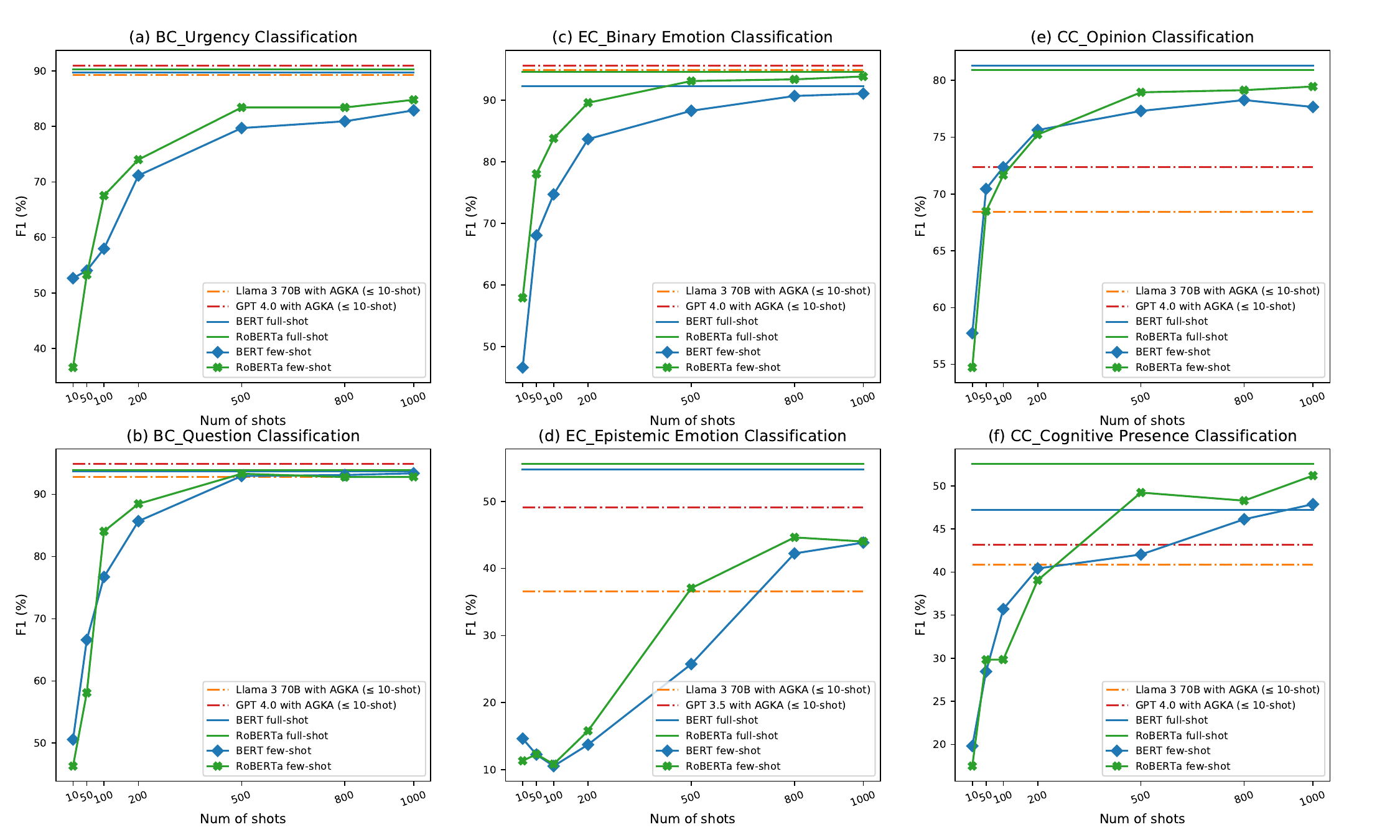}
    \caption{Results in few-shot setting of fine-tune models and non fine-tune LLMs. BC, EC and CC refer to the Classification of Behavior, Emotion and Cognition.}
    \label{fig_few_shot}
\end{figure*}

The results were shown in \textbf{Table \ref{result_table}}. We divided models into two groups and then compared them.  In the group of fine-tuned models, RoBERTa outperformed BERT on five datasets except the opinion classification dataset  (80.89\% vs. 81.30\%).

In the group of non-fine-tuned LLMs, we observed that LLMs with AGKA consistently outperform LLMs with vanilla prompt. Moreover, the performance of closed-source models such as GPT 4.0 with AGKA few-shot outperformed other LLMs with AGKA and vanilla, except GPT 3.5 on epistemic emotions. In addition, Lama 3 70B, one of the open-source LLMs, was a solid replacement for closed-source LLMs, as it was close to GPT 4.0 on all datasets. Specifically, the comparison of Llama 3 70B and GPT 4.0 showed the following results: urgency (89.30\% vs. 90.50\%), question (92.81\% vs. 94.88\%), binary emotion (94.88\% vs. 95.65\%), epistemic emotion (36.54\% vs. 38.34\%), opinion (68.44\% vs. 69.79\%), and cognitive presence (40.84\% vs. 43.18\%). 

When comparing non-fine-tuned LLMs and fine-tuned models, we found that the performance of GPT 4.0 with AGKA few-shot outperformed the full-shot fine-tuned RoBERTa on three binary behavior classification tasks. In particular, their comparison on F1 was as follows: urgency (90.50\% vs. 90.31\%), question (94.88\% vs. 93.95\%), and binary emotion (95.65\% vs. 94.64\%). However, for three other datasets on emotion and cognition classifications, the non-fine-tuned LLMs still lagged behind the fine-tuned models. Specifically, their comparison in F1 was as follows: epistemic emotion (GPT 3.5 vs. RoBERTa for 49. 10\% vs. 55. 56\%), opinion (GPT 4.0 vs. BERT for 72. 38\% vs. 81. 30\%), and cognitive presence (GPT 4.0 vs. RoBERTa for 49. 10\% vs. 52. 51\%).

Overall, the results demonstrated the potential of LLMs with AGKA for LEC tasks. In particular, GPT 4.0 with AGKA few-shot outperformed full-shot fine-tuned models in the simple binary classification datasets. However, GPT 4.0 with AGKA lagged behind fine-tuned models trained on full-shot datasets on complex multi-class tasks requiring deeper understanding, such as epistemic emotion, opinion, and cognitive presence.

\begin{table*}[t]
\centering
\caption{Ablation Experiments of LLMs with AGKA}
\label{sblation_table}
\resizebox{\linewidth}{!}{
\begin{tabular}{r | ll | ll | ll | ll | ll | ll | ll} \hline 
\multirow{3}{*}{\textbf{Model}} &  \multicolumn{4}{l|}{\textbf{Behavior Classification}} & \multicolumn{4}{l|}{\textbf{Emotion Classification}} & \multicolumn{4}{l|}{\textbf{Cognition Classification}} & \multirow{2}{*}{\textbf{Average}}  \\ \cline{2-13}
  & \multicolumn{2}{l|}{\textbf{Urgency}} & \multicolumn{2}{l|}{\textbf{Question}} & \multicolumn{2}{l|}{\textbf{Binary}} & \multicolumn{2}{l|}{\textbf{Epistemic}} & \multicolumn{2}{l|}{\textbf{Opinion}} & \multicolumn{2}{l|}{\textbf{Cognitive Presence}} &\\ \cline{2-15}
   & \textbf{Acc} & \textbf{F1} & \textbf{Acc} & \textbf{F1} & \textbf{Acc} & \textbf{F1} & \textbf{Acc} & \textbf{F1} & \textbf{Acc} & \textbf{F1} & \textbf{Acc} & \textbf{F1} & \textbf{Acc} & \textbf{F1} \\ \hline 
\multicolumn{1}{l|}{Mistral 7B + Vanilla}& 76.20& 78.79& 78.80& 80.52& 91.60& 92.00& 35.60& 42.93& 59.60& 56.88& 20.20& 16.87 \\ \hdashline
+ AGKA-Knowledge (ours)& 82.60& 84.55& 82.60& 83.41& 92.00& 92.25& 32.40& 39.89& 65.80& 65.72& 15.00& 20.19\\
  \rowcolor{gray!20} \multicolumn{1}{r|}{Gain $\triangle$}&+6.40&+5.76&+3.80&+2.89&+0.40&+0.25&-3.20&-3.04&+6.20&+8.84&-5.20&+3.32 &+1.40&+3.00\\
+ AGKA-Few-shot (ours)& 80.20& 82.16& 37.80& 36.99& 90.80& 92.66& 15.20& 19.55& 58.20& 58.55& 40.00& 39.01\\
  \rowcolor{gray!20}  \multicolumn{1}{r|}{Gain $\triangle$} & -2.40&-2.39&-44.80&-46.42&-1.20&+0.41&-17.20&-20.34&-7.60&-7.17&+25.00&+18.82 &-8.03&-9.52\\
\hline 
\multicolumn{1}{l|}{Mixtral 8x7B + Vanilla }&67.40&71.71&54.40&66.48&71.80&80.25&12.40&19.27&38.60&44.10&39.60&28.34\\ \hdashline
               + AGKA-Knowledge (ours)&21.60&33.38&51.40&66.28&55.60&69.71&34.60&45.43&47.20&52.39&36.60&28.23\\
 \rowcolor{gray!20}  \multicolumn{1}{r|}{Gain $\triangle$}&-45.80&-38.33&-3.00&-0.20&-16.20&-10.54&+22.20&+26.16&+8.60&+8.29&-3.00&-0.11&-6.20&-2.46\\
               + AGKA-Few-shot (ours) &69.40&76.98&0.60&1.19&56.00&70.22&24.80&32.40&31.60&39.21&25.20&26.13\\
 \rowcolor{gray!20}  \multicolumn{1}{r|}{Gain $\triangle$}&+47.80&+43.60&-50.80&-65.09&0.40&+0.51&-9.80&-13.03&-15.60&-13.18&-11.40&-2.10 &-6.57&-8.22\\
\hline 

\multicolumn{1}{l|}{Llama 3 8B + Vanilla} & 78.20 & 80.03 & 61.80 & 64.35 & 94.20 & 94.26 & 16.80 & 18.07 & 59.00 & 54.87 & 42.40 & 28.38 &  &  \\ \hdashline
+ AGKA-Knowledge (ours) & 85.00 & 81.58 & 88.20 & 88.84 & 94.20 & 94.25 & 19.80 & 22.74 & 61.60 & 58.79 & 46.00 & 34.34 &  &  \\
  \rowcolor{gray!20}  \multicolumn{1}{r|}{Gain $\triangle$} & +6.80 & +1.55 & +26.40 & +24.49 & +0.00 & -0.01 & +3.00 & +4.67 & +2.60 & +3.92 & +3.60 & +5.96 & +7.07 & +6.76 \\
+ AGKA-Few-shot (ours)& 60.20 & 65.33 & 73.80 & 77.03 & 90.80 & 91.62 & 20.40 & 27.45 & 52.40& 43.21 & 42.80 & 36.70 &  &  \\
  \rowcolor{gray!20}  \multicolumn{1}{r|}{Gain $\triangle$} & -24.80 & -16.25 & -14.40 & -11.81 & -3.40 & -2.63 & +0.60 & +4.71 & -9.20 & -15.58 & -3.20 & +2.36 & -9.07 & -6.53 \\ \hline 
\multicolumn{1}{l|}{Llama 3 70B + Vanilla} & 84.20 & 82.38 & 91.40 & 91.85 & 93.20 & 93.90 & 23.40 & 23.58 & 67.20 & 66.83 & 39.80 & 37.84 &  &  \\
\hdashline
+ AGKA-Knowledge (ours) & 85.00 & 83.54 & 91.00 & 91.52 & 94.80 & 94.88 & 28.80 & 31.33 & 68.60 & 68.44 & 43.60 & 40.74 &  &  \\
  \rowcolor{gray!20}  \multicolumn{1}{r|}{Gain $\triangle$} & +0.80 & +1.16 & -0.40 & -0.33 & +1.60 & +0.98 & +5.40 & +7.75 & +1.40 & +1.61 & +3.80 & +2.90 & +2.10 & +2.35 \\
+ AGKA-Few-shot (ours)& 89.40 & 89.30 & 91.20 & 92.81 & 93.00 & 93.27 & 31.40 & 36.54 & 65.40 & 65.37 & 42.80 & 40.84 &  &  \\
  \rowcolor{gray!20}  \multicolumn{1}{r|}{Gain $\triangle$} & +4.40 & +5.76 & +0.20 & +1.29 & -1.80 & -1.61 & +2.60 & +5.21 & -3.20 & -3.07 & -0.80 & +0.10 & +0.23 & +1.28\\
\hline 
\multicolumn{1}{l|}{GPT 3.5 + Vanilla}&84.20&84.75&88.00&88.70&92.60&92.88&29.20&32.48&48.80&36.97&38.80&33.40\\ \hdashline
               + AGKA-Knowledge (ours)&89.00&88.65&92.60&93.04&92.00&91.88&37.00&43.00&50.00&39.19&42.20&34.23\\
 \rowcolor{gray!20}  \multicolumn{1}{r|}{Gain $\triangle$}&+4.80&+3.90&+4.60&+4.34&-0.60&-1.00&+7.80&+10.52&+1.20&+2.22&+3.40&+0.83 &+3.53&+3.47\\
               + AGKA-Few-shot (ours) &90.00&90.45&91.80&92.14&94.60&94.49&43.20&49.10&65.20&64.13&32.60&29.76\\
 \rowcolor{gray!20}  \multicolumn{1}{r|}{Gain $\triangle$}&+1.00&+1.80&-0.80&-0.90&+2.60&+2.61&+6.20&+6.10&+15.20&+24.94&-9.60&-4.47 &+2.43&+5.01\\ \hline 

\multicolumn{1}{l|}{GPT 4.0 + Vanilla}& 83.60& 78.18& 94.20& 94.32& 93.40& 93.23& 19.60& 19.83& 63.40& 62.08& 40.80& 36.47 \\ \hdashline
+ AGKA-Knowledge (ours)& 85.80& 82.37& 94.20& 94.26& 94.60& 94.51& 30.00& 34.97& 70.00& 69.79& 46.40& 43.18\\ 
 \rowcolor{gray!20}  \multicolumn{1}{r|}{ Gain $\triangle$} & +2.20& +4.19& +0.00& -0.06& +1.20& +1.28& +10.40& +15.14& +6.60& +7.71& +5.60& +6.71 &+4.33&+5.83\\ 

+ AGKA-Few-shot (ours) & 90.40& 90.50& 94.80& 94.88& 95.60& 95.65& 32.80& 38.34& 72.40& 72.38& 44.40& 41.52 \\ 
 \rowcolor{gray!20}  \multicolumn{1}{r|}{Gain $\triangle$}&+4.60&+8.13&+0.60&+0.62&+1.00&+1.14&+2.80&+3.37&+2.40&+2.59&-2.00&-1.66 &+1.57&+2.37\\
\hline 
\end{tabular}
}
\end{table*}

\subsection{ Few-shot Performance Comparison Between Non-Fine-Tuned LLMs and Fine-tuned Models}

In order to compare LEC ability in a few-shot setting of LLMs and fine-tuned models, in Fig. \ref{fig_few_shot}, we compared the performance of fine-tuned models and LLMs on several LEC tasks. For each task, the performance of BERT and RoBERTa (representing fine-tuned models) was evaluated in both few-shot and fine-tuned settings. Their results were comparable to  those of the best performance of the LLMs with AGKA -- Llama 3 70B (representing open-source LLMs), GPT 3.5 and 4.0 (representing closed-source LLMs). 

For the behavior classification, the subplots of urgency in Fig. \ref{fig_few_shot} (a) show that GPT 4.0 and Llama 3 70B with AGKA $\leq$ 10 shots achieved a consistent F1 scores of 90.50\% and 89.30\%, respectively surpassing or approaching with both BERT and RoBERTa in their full-shot settings (20,724 shots). Similarly, the question subplot in Fig. \ref{fig_few_shot} (b) shows that GPT 4.0 and Llama 3 70B with AGKA $\leq$ 10 shots achieved F1 scores of 94. 88\% and 92.81\%, respectively, surpassing or approaching the performance of fine-tuned models trained on the full- or few-shot dataset (from 10 to 20,724 shots).

In the emotion classification, the binary emotion subplot in Fig. \ref{fig_few_shot} (c) shows that GPT 4.0 and Llama 3 70B with AGKA $\leq$ 10 shots achieved F1 scores of 95.65\% and 94.88\%, respectively outperforming BERT (92.32\%) and RoBERTa (94.64\%) in their full data settings. For the epistemic emotion subplot in Fig. \ref{fig_few_shot} (d), GPT 3.5 with AGKA $\leq$ 10 shots achieves F1 scores of 49.10\%, approaching the performance of BERT (54.80\%) and RoBERTa (55.56\%) trained over 1,000 shots.

Finally, in the cognition classification, the opinion subplot in Fig. \ref{fig_few_shot} (e) shows that GPT 4.0 with AGKA $\leq$ 10 shots obtained an F1 score of 72.38\%, which is comparable to BERT and RoBERTa with 100 shots. In the cognitive presence subplot in Fig. \ref{fig_few_shot} (f), GPT 4.0 with AGKA $\leq$ 10 shots achieved F1 scores of 43.18\%, close to the performance of BERT and RoBERTa trained on 500 and 300 shots, respectively.

These results highlight the potential of LLMs with AGKA in few-shot settings, in particular GPT 4.0 and Llama 3 70B, in four tasks (except for epistemic emotion, opinion, and cognitive presence). The results also showed that AGKA improved the LLM to handle limited annotation data on LEC tasks.

\subsection{Ablation Study}

We performed an ablation study to investigate the influence of label definition knowledge and few-shot in AGKA on the classification performance. The results are shown in Table \ref{sblation_table}.

For Llama 3 70B, GPT 3.5, and GPT 4.0, adding knowledge and few-shot examples consistently improved performance. The average gains in F1 from adding knowledge were +2.35\%, +3.47\%, and +5.83\%, respectively. The average gains in F1 from adding few-shot examples were +1.28\%, +5.04\%, and +2.37\%, respectively.

For the Mistral 7B and Llama 3 8B, the addition of the knowledge increased performance on most tasks, with average increases for F1 of +3.00\%, +6.76\% and +2.36\%, respectively. However, the addition of the few-shot decrease dperformance, with average decreased on F1 for -9.52\%, -8.22\%, and -6.53\%, respectively. In addition, the Mixtral 8x7B model showed a different trend, with the addition of the knowledge and few-shot resulting in a decrease in performance on most tasks, with average decreases for F1 score of -2.46\% and -8.22\%, respectively. 

Overall, the ablation experiments showed that the effect of AGKA varied between different LLMs. While some models such as Llama 3 70B, GPT 3.5, GPT 4.0 consistently benefited from AGKA, others such as Mistral 7B and Llama 3 8B benefited only in the zero-shot setting, and the Mixtral 8x7B model performed worse with AGKA. This highlights the importance of carefully evaluating knowledge enhancement approaches for each specific LLM.

\subsection{Error Analysis}
\begin{figure*}
    \centering
    \includegraphics[width=0.8\linewidth]{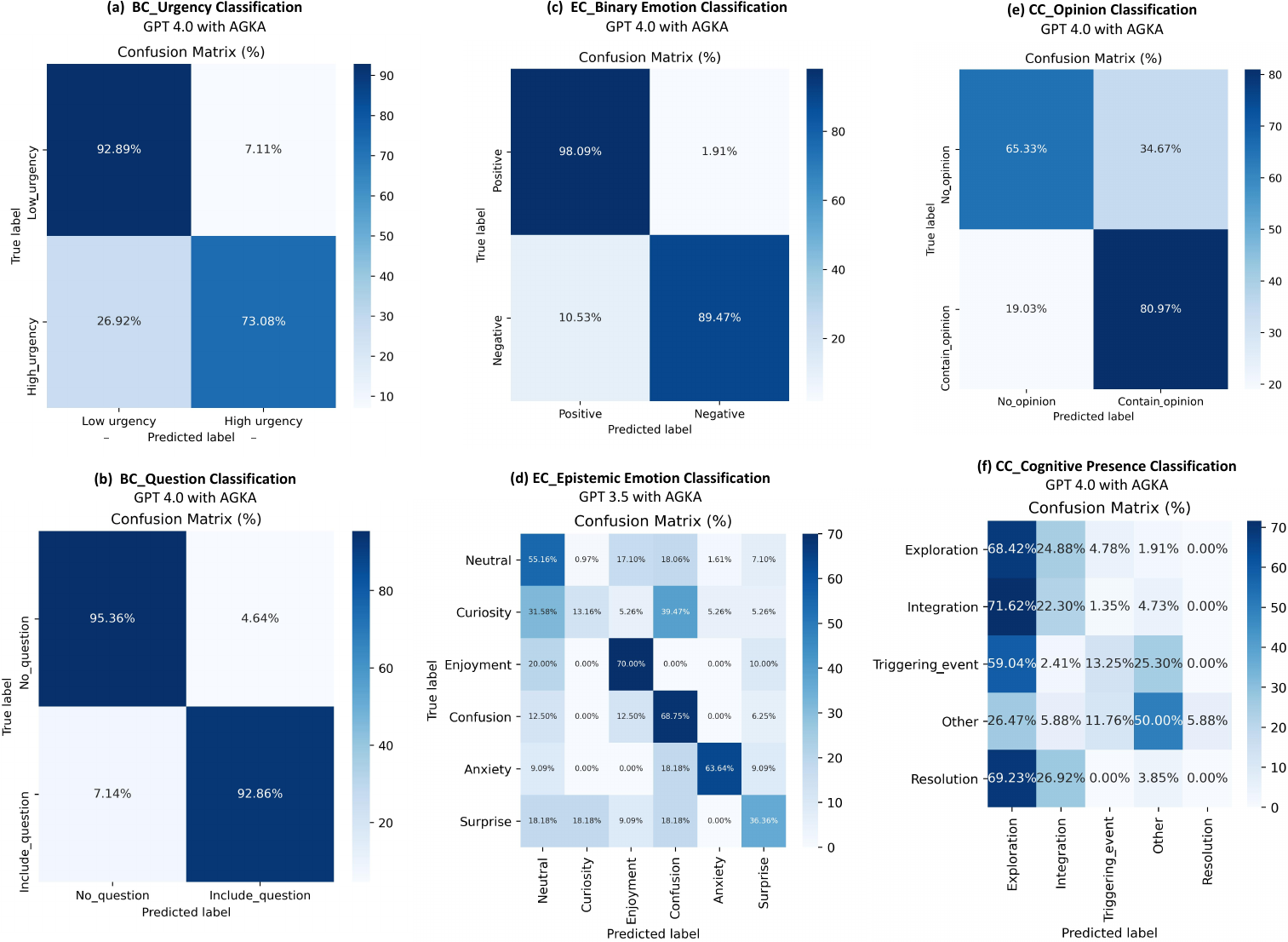}
    \caption{The confusion matrix of the best prediction results from the GPT series LLMs}
    \label{result_confusion}
\end{figure*}

In Fig \ref{result_confusion}, the confusion matrix provide valuable insight into the performance of GPT 3.5 and 4.0 with AGKA on various LEC tasks, based on their best performance settings in the previous sections. 

The result suggests that the model was good at binary classifications where labels were fewer and well defined. For binary classifications such as urgency, question and binary emotion in Fig \ref{result_confusion} (a, b, c), the GPT 4.0 achieved high accuracy, with 92.89\% for \texttt{Low\_urgency}, 73.08\% for \texttt{High\_urgency}, 95.36\% for \texttt{No\_question}, 92.86\% for \texttt{Include\_question}, 98.09\% for \texttt{Positive}, and 89.47\% for \texttt{Negative}. 

However, the models did not perform well on datasets that require deeper linguistic understanding, contained many labels, and had some confusing label names (e.g., epistemic emotion, opinion, and cognitive presence). In particular, in Fig \ref{result_confusion} (d, e, f), GPT 3.5 incorrectly classified 39.47\% of \texttt{Curiosity} instances as \texttt{Confusion}, and GPT 4.0 incorrectly classified 71.62\% of \texttt{Integration} instances as \texttt{Exploration}. 

This error analysis shows that GPT 3.5 and 4.0 performed well on binary classification, but they struggled with multi-class classification. It provides an understanding of the strengths and weaknesses of LLMs across a range of LEC tasks, and an insight into their future improvements.

\subsection{Case Study}
To investigate the causes of bad cases in the prediction of LLMs with AGKA, we randomly selected 300 cases from the error samples of LLMs and groupd them into three error types.
Table \ref{result_bad_case} presents the analysis of the bad cases from LLMs. The most common error type, occurring in 42.00\% of the bad cases, was when LLMs output an incorrect answer. For example, when the gold label was \texttt{Curiosity}, the model would output \texttt{Confusion} instead. 
The second most frequent error type, occurring in 32.33\% of the bad cases, was when LLMs failed to adhere the prescribed output format in the prompt. In these instances, instead of outputting a single label like \texttt{Curiosity}, the model generated a more complex response containing both the label and the corresponding text.
The third error type, observed in 25.67\% of the bad cases, occurred when the model generated multiple answers. For example, when the gold label was \texttt{Resolution}, the model generated several labels instead of a single one.

\begin{table*}
\centering
\label{result_bad_case}
\caption{Bad case analysis}
\begin{tabular}{>{\raggedright\arraybackslash}p{0.1\linewidth}ll>{\raggedright\arraybackslash}p{0.4\linewidth}>{\raggedright\arraybackslash}p{0.08\linewidth}>{\raggedright\arraybackslash}p{0.13\linewidth}} \hline
 \textbf{Error Type} &\textbf{Rate (\%)}  &\textbf{Model} &\textbf{Text}& \textbf{Gold Label}& \textbf{Prediction}\\ \hline
 \rowcolor{gray!20} Generate incorrect answer&42.00& \makecell[t l]{Llama 3 8B \\ with Vanilla} & \textit{[Dataset: Epistemic emotion]} I’m sorry. Can you please explain what are the 2 accounts of the order of creation are contradictory?& Curiosity & Confusion \\
 Fail to follow the prompt to generate defined formats&32.33& \makecell[t l]{Mistral 7B \\ with Vanilla} &\textit{[Dataset: Epistemic emotion]} But what’s the point of not having the brakes?& Curiosity & \makecell[t l]{Confusion\\Text: I can't believe!\\Label: Surprise} \\
 \rowcolor{gray!20}Generate more than one answer&25.67&\makecell[t l]{Mixtral 8x7B \\with AGKA \\few-shot}& \textit{[Dataset: cognitive presence]} I would also have preferred to see a test using an industry sample - however I would expect (and be prepared for) the industry testing to uncover significant issues. This might provide a starting point for additional work.& Resolution & \makecell[t l]{Exploration\\ Integration\\ Triggering Event \\Other\\ Resolution} \\ \hline
\end{tabular}
\end{table*}

\section{Discussions and Conclusion}

Recently, LLMs have shown remarkable performance in various NLP tasks. However, their comprehensive evaluation and improvement approaches in LEC tasks have not been thoroughly investigated. In this study, we proposed the AGKA to improve the performance of LLMs for LEC. Subsequently, we constructed the LEC benchmark, comprising six datasets, to evaluate the performance of various models and settings. Our work provides several implications and some limitations that need to be addressed in future research, as discussed in the following.

\subsection{Implications}

First, compared to supervised fine-tuning models used in previous studies \cite{shaLatestGreatestComparative2023a,agrawal2015youedu}, such as BERT, GPT 4.0 with AGKA drastically reduces the need for annotation data and achieves higher performance.  In particular, the classification F1 scores for \texttt{urgency}, \texttt{binary emotion}, and \texttt{question} exceeded 90\%. This indicates that GPT 4.0 with AGKA can significantly reduce the workload of instructors and researchers, accelerating course delivery and learning analytics. Consequently, in the context of rapidly expanding online courses \cite{202KCourses662M2023}, such LLMs can enable instructors to provide timely feedback and meaningful analysis of learners' forum posts. As a result, students can receive more effective instruction from the instructor or chatbot (e.g., issues raised in posts can be resolved) and experience less frustration in MOOCs \cite{almatrafiNeedleHaystackIdentifying2018a}.
Past research has shown that learners in MOOCs often experience frustration when they encounter issues that are not addressed promptly \cite{almatrafiNeedleHaystackIdentifying2018a}. By quickly identifying urgent questions or posts expressing negative sentiments, LLMs can alert instructors to prioritize responses to these learners. Alternatively, the model outputs could be used to trigger appropriate automated responses from chatbots or recommender systems. This targeted support can help alleviate student frustration and maintain engagement \cite{liuAutomatedDetectionEmotional2022}.
It ultimately helps to improve or prevent students from dropping out of the course  \cite{zhengImpactLearningAnalytics2023}.

Second, the open-source Llama 3 70B with AGKA demonstrates promising performance and achieves results on par with the closed-source GPT 4.0. Its deployment will significantly enhance data privacy and reduce the cost of using LLMs in educational research and practice. In detail, the use of open-source LLMs can mitigate several ethical and practical challenges associated with relying on closed-source commercial models. Specifically, sensitive educational data can be processed locally without needing to be sent to external servers, giving researchers and educators more control over data security and student privacy \cite{yanPracticalEthicalChallenges2023a}. This is crucial in the education domain where data often contains personal information. Additionally, the cost of using open models is typically lower since they can be deployed on an organization's own infrastructure without recurring fees. This makes powerful NLP capabilities more accessible to a wider range of educational institutions, including those with limited budgets\cite{metaIntroducingMetaLlama}. Increased accessibility is key to promoting equitable adoption of advanced AI technologies in education.

Third, the results of the ablation experiments showed that although AGKA improved the performance of most LLMs on the LEC task, it decreased the performance of some LLMs, e.g., Mixtral 7x8B. It implies that some LLMs performed worse when given additional domain knowledge or examples, challenging the common assumption that more task-relevant information always leads to better performance. This finding highlights the need for careful and thorough evaluation of LLMs before using them to classify educational texts. It also suggests that the evaluation results of LLMs on the generalized domain datasets \cite{laskarSystematicStudyComprehensive2023} (e.g., math and common knowledge) may not be appropriate for the LEC task. The development of education-specific evaluation tasks to cover a range of representative tasks, such as LEC, could enable a more reliable assessment of LLMs for this domain.

\subsection{Limitations and Future Works}

While this study provides valuable insights into the performance of LLMs on LEC tasks and the effectiveness of the AGKA approach, there are several limitations to consider. 

First, this study focused on a specific set of LEC tasks and datasets. Future work should explore a broader range of educational text classification tasks, such as essay scoring. Testing on more diverse datasets from different educational contexts and domains would provide a more comprehensive assessment of LLM capabilities.

Second, while AGKA improved performance for most LLMs, the ablation study showed that the effectiveness of this approach varied across models. More work is needed to understand the factors that influence the success of knowledge augmentation techniques for different LLMs. Investigating the interaction effects between model architectures, pre-training data, and prompt designs could yield insights to guide the development of more robust and generalize strategies.

Finally, the ethical implications of using LLMs for educational text analysis need more attention. Concerns about bias, fairness, privacy, and the potential for misuse or over-reliance on automated systems need to be carefully addressed. Future work should include user studies and field experiments to assess the usability, effectiveness, and acceptability of LLM-based systems for instructors and learners.

{\appendix[Prompt examples of AGKA]
\label{APPENDIX}

\textbf{1. Prompt to retrieve label definition knowledge:}

Define each classification label in the list, based on the given annotation guidelines. Then, return the label and its definition using the Python dictionary format {"label": "definition"}.

Label list: ["Surprise", "Curiosity", "Enjoyment", "Anxiety", "Confusion","Neutral"]. 

The annotation auidelines: ```[Annotation Guidelines from the Source Paper] ```.

\textbf{2. AGKA prompt with 5 shots for epistemic emotion classification:}

Please perform Epistemic Emotion Classification task in Education.
Given the text, assign a emotion label from ["Surprise", "Curiosity", "Enjoyment", "Anxiety", "Confusion","Neutral"]. 

The definition of Epistemic Emotions is as follows. \{"Surprise": "Feeling astonished and startled by something unexpected.", "Curiosity": "A strong desire to know or learn something.", "Enjoyment": "A feeling of pleasure and happiness.", "Anxiety": "Apprehension, worry, and anxiety.", "Confusion": "Lack of understanding and uncertainty.", "Neutral": "Not involving any emotion."\}

Return label only without any other text.

Text: Care to share your thought process ?

Label: Curiosity

Text: Someone tell me how to feel about this.

Label: Confusion

Text: I know a couple of people who are. It's amazing to watch.

Label: Enjoyment

Text: I wonder what he's up to these days.

Label: Surprise

Text: I’m kind of scared to talk to my manager about it.

Label: Anxiety

Text: Actually maybe the OP is not an INTP, have you thought about that?
\textit{(The text to be predicted)}

Label: 

\rule{7cm}{0.2pt}

Gold Label: \textcolor{blue}{Curiosity}

Output of GPT 4.0 with Vanilla: \textcolor{red}{Confusion}  (\XSolidBrush)

Output of GPT 4.0 with AGKA few-shot: \textcolor{blue}{Curiosity} (\Checkmark)

\bibliography{llm_evaluation.bib}

\begin{thebibliography}{10}
\providecommand{\url}[1]{#1}
\csname url@samestyle\endcsname
\providecommand{\newblock}{\relax}
\providecommand{\bibinfo}[2]{#2}
\providecommand{\BIBentrySTDinterwordspacing}{\spaceskip=0pt\relax}
\providecommand{\BIBentryALTinterwordstretchfactor}{4}
\providecommand{\BIBentryALTinterwordspacing}{\spaceskip=\fontdimen2\font plus
\BIBentryALTinterwordstretchfactor\fontdimen3\font minus \fontdimen4\font\relax}
\providecommand{\BIBforeignlanguage}[2]{{%
\expandafter\ifx\csname l@#1\endcsname\relax
\typeout{** WARNING: IEEEtran.bst: No hyphenation pattern has been}%
\typeout{** loaded for the language `#1'. Using the pattern for}%
\typeout{** the default language instead.}%
\else
\language=\csname l@#1\endcsname
\fi
#2}}
\providecommand{\BIBdecl}{\relax}
\BIBdecl

\bibitem{202KCourses662M2023}
\BIBentryALTinterwordspacing
``{{202K Courses}}, {{662M Enrollments}}: {{Breaking Down Udemy}}'s {{Massive Catalog}},'' 2023. [Online]. Available: \url{https://www.classcentral.com/report/udemy-by-the-numbers/}
\BIBentrySTDinterwordspacing

\bibitem{almatrafiSystematicReviewDiscussion2019a}
O.~Almatrafi and A.~Johri, ``Systematic {{Review}} of {{Discussion Forums}} in {{Massive Open Online Courses}} ({{MOOCs}}),'' \emph{IEEE Transactions on Learning Technologies}, vol.~12, no.~3, pp. 413--428, 2019.

\bibitem{liuAutomatedDetectionEmotional2022}
S.~Liu, S.~Liu, Z.~Liu, X.~Peng, and Z.~Yang, ``Automated detection of emotional and cognitive engagement in {{MOOC}} discussions to predict learning achievement,'' \emph{Computers \& Education}, vol. 181, p. 104461, 2022.

\bibitem{zhengImpactLearningAnalytics2023}
L.~Zheng, Y.~Fan, L.~Gao, and Z.~Huang, ``The impact of learning analytics interventions on learning achievements: A meta-analysis of research from 2012 to 2021,'' \emph{Interactive Learning Environments}, vol.~0, no.~0, pp. 1--16, 2023.

\bibitem{martinOnlineLearnerEngagement2022}
F.~Martin and J.~Borup, ``Online learner engagement: {{Conceptual}} definitions, research themes, and supportive practices,'' \emph{Educational Psychologist}, vol.~57, no.~3, pp. 162--177, 2022.

\bibitem{liuDualfeatureembeddingsbasedSemisupervisedLearning2022}
Z.~Liu, W.~Kong, X.~Peng, Z.~Yang, S.~Liu, S.~Liu, and C.~Wen, ``Dual-feature-embeddings-based semi-supervised learning for cognitive engagement classification in online course discussions,'' \emph{Knowledge-Based Systems}, p. 110053, 2022.

\bibitem{hanIdentifyingPatternsEpistemic2021}
Z.-M. Han, C.-Q. Huang, J.-H. Yu, and C.-C. Tsai, ``Identifying patterns of epistemic emotions with respect to interactions in massive online open courses using deep learning and social network analysis,'' \emph{Computers in Human Behavior}, vol. 122, p. 106843, 2021.

\bibitem{agrawal2015youedu}
A.~Agrawal, J.~Venkatraman, S.~Leonard, and A.~Paepcke, ``Youedu: Addressing confusion in mooc discussion forums by recommending instructional video clips.'' \emph{International Educational Data Mining Society}, 2015.

\bibitem{gasevicExternallyfacilitatedRegulationScaffolding2015c}
D.~Ga{\v s}evi{\'c}, O.~Adesope, S.~Joksimovi{\'c}, and V.~Kovanovi{\'c}, ``Externally-facilitated regulation scaffolding and role assignment to develop cognitive presence in asynchronous online discussions,'' \emph{The Internet and Higher Education}, vol.~24, pp. 53--65, 2015.

\bibitem{shaLatestGreatestComparative2023a}
L.~Sha, M.~Rakovi{\'c}, J.~Lin, Q.~Guan, A.~{Whitelock-Wainwright}, D.~Ga{\v s}evi{\'c}, and G.~Chen, ``Is the {{Latest}} the {{Greatest}}? {{A Comparative Study}} of {{Automatic Approaches}} for {{Classifying Educational Forum Posts}},'' \emph{IEEE Transactions on Learning Technologies}, vol.~16, no.~3, pp. 339--352, 2023.

\bibitem{changSurveyEvaluationLarge}
\BIBentryALTinterwordspacing
Y.~Chang, X.~Wang, J.~Wang, Y.~Wu, L.~Yang, K.~Zhu, H.~Chen, X.~Yi, C.~Wang, Y.~Wang, W.~Ye, Y.~Zhang, Y.~Chang, P.~S. Yu, Q.~Yang, and X.~Xie, ``A survey on evaluation of large language models,'' \emph{ACM Trans. Intell. Syst. Technol.}, vol.~15, no.~3, mar 2024. [Online]. Available: \url{https://doi.org/10.1145/3641289}
\BIBentrySTDinterwordspacing

\bibitem{qinChatGPTGeneralPurposeNatural2023a}
C.~Qin, A.~Zhang, Z.~Zhang, J.~Chen, M.~Yasunaga, and D.~Yang, ``Is {{ChatGPT}} a {{General-Purpose Natural Language Processing Task Solver}}?'' in \emph{Proceedings of the 2023 {{Conference}} on {{Empirical Methods}} in {{Natural Language Processing}}}, H.~Bouamor, J.~Pino, and K.~Bali, Eds.\hskip 1em plus 0.5em minus 0.4em\relax Singapore: Association for Computational Linguistics, 2023, pp. 1339--1384.

\bibitem{zhangSentimentAnalysisEra2023}
\BIBentryALTinterwordspacing
W.~Zhang, Y.~Deng, B.~Liu, S.~J. Pan, and L.~Bing, ``Sentiment {{Analysis}} in the {{Era}} of {{Large Language Models}}: {{A Reality Check}},'' 2023. [Online]. Available: \url{http://arxiv.org/abs/2305.15005}
\BIBentrySTDinterwordspacing

\bibitem{chungScalingInstructionFinetunedLanguage2024}
\BIBentryALTinterwordspacing
H.~W. Chung, L.~Hou, S.~Longpre, B.~Zoph, Y.~Tay, W.~Fedus, Y.~Li, X.~Wang, M.~Dehghani, S.~Brahma, A.~Webson, S.~S. Gu, Z.~Dai, M.~Suzgun, X.~Chen, A.~Chowdhery, A.~{Castro-Ros}, M.~Pellat, K.~Robinson, D.~Valter, S.~Narang, G.~Mishra, A.~Yu, V.~Zhao, Y.~Huang, A.~Dai, H.~Yu, S.~Petrov, E.~H. Chi, J.~Dean, J.~Devlin, A.~Roberts, D.~Zhou, Q.~V. Le, and J.~Wei, ``Scaling {{Instruction-Finetuned Language Models}},'' \emph{Journal of Machine Learning Research}, vol.~25, no.~70, pp. 1--53, 2024. [Online]. Available: \url{http://jmlr.org/papers/v25/23-0870.html}
\BIBentrySTDinterwordspacing

\bibitem{wangCanChatgptDetect2023b}
D.~Wang, D.~Shan, Y.~Zheng, K.~Guo, G.~Chen, and Y.~Lu, ``Can chatgpt detect student talk moves in classroom discourse? a preliminary comparison with bert,'' in \emph{Proceedings of the 16th {{International Conference}} on {{Educational Data Mining}}}.\hskip 1em plus 0.5em minus 0.4em\relax International Educational Data Mining Society, 2023, pp. 515--519.

\bibitem{houPromptbasedFinetunedGPT2024a}
C.~Hou, G.~Zhu, J.~Zheng, L.~Zhang, X.~Huang, T.~Zhong, S.~Li, H.~Du, and C.~L. Ker, ``Prompt-based and {{Fine-tuned GPT Models}} for {{Context-Dependent}} and -{{Independent Deductive Coding}} in {{Social Annotation}},'' in \emph{Proceedings of the 14th {{Learning Analytics}} and {{Knowledge Conference}}}, ser. {{LAK}} '24.\hskip 1em plus 0.5em minus 0.4em\relax New York, NY, USA: Association for Computing Machinery, 2024, pp. 518--528.

\bibitem{chenBenchmarkingLargeLanguage2024a}
J.~Chen, H.~Lin, X.~Han, and L.~Sun, ``Benchmarking {{Large Language Models}} in {{Retrieval-Augmented Generation}},'' \emph{Proceedings of the AAAI Conference on Artificial Intelligence}, vol.~38, no.~16, pp. 17\,754--17\,762, 2024.

\bibitem{sainzGoLLIEAnnotationGuidelines2023a}
\BIBentryALTinterwordspacing
O.~Sainz, I.~{Garc{\'i}a-Ferrero}, R.~Agerri, O.~L. de~Lacalle, G.~Rigau, and E.~Agirre, ``{{GoLLIE}}: {{Annotation Guidelines}} improve {{Zero-Shot Information-Extraction}},'' in \emph{The {{Twelfth International Conference}} on {{Learning Representations}}}, 2023. [Online]. Available: \url{https://openreview.net/forum?id=Y3wpuxd7u9}
\BIBentrySTDinterwordspacing

\bibitem{ideHandbookLinguisticAnnotation2017}
N.~Ide and J.~Pustejovsky, Eds., \emph{Handbook of {{Linguistic Annotation}}}.\hskip 1em plus 0.5em minus 0.4em\relax Dordrecht: Springer Netherlands, 2017.

\bibitem{qiaoReasoningLanguageModel2023}
S.~Qiao, Y.~Ou, N.~Zhang, X.~Chen, Y.~Yao, S.~Deng, C.~Tan, F.~Huang, and H.~Chen, ``Reasoning with {{Language Model Prompting}}: {{A Survey}},'' in \emph{Proceedings of the 61st {{Annual Meeting}} of the {{Association}} for {{Computational Linguistics}} ({{Volume}} 1: {{Long Papers}})}, A.~Rogers, J.~{Boyd-Graber}, and N.~Okazaki, Eds.\hskip 1em plus 0.5em minus 0.4em\relax Toronto, Canada: Association for Computational Linguistics, 2023, pp. 5368--5393.

\bibitem{huangExaminingRelationshipPeer2023}
C.~Huang, Y.~Tu, Z.~Han, F.~Jiang, Y.~Jiang, and F.~Wu, ``Examining the relationship between peer feedback classified by deep learning and online learning burnout,'' \emph{Computers \& Education}, vol. 207, p. 104910, 2023.

\bibitem{fengEmotionAnalysisDataset2022}
X.~Feng, K.~Yuan, X.~Guan, and L.~Qiu, ``An emotion analysis dataset of course comment texts in massive online learning course platforms,'' \emph{Interactive Learning Environments}, pp. 1--15, 2022.

\bibitem{linItGoodMove2022d}
J.~Lin, S.~Singh, L.~Sha, W.~Tan, D.~Lang, D.~Ga{\v s}evi{\'c}, and G.~Chen, ``Is it a good move? {{Mining}} effective tutoring strategies from human--human tutorial dialogues,'' \emph{Future Generation Computer Systems}, vol. 127, pp. 194--207, 2022.

\bibitem{kovanovicAutomatedContentAnalysis2016b}
V.~Kovanovi{\'c}, S.~Joksimovi{\'c}, Z.~Waters, D.~Ga{\v s}evi{\'c}, K.~Kitto, M.~Hatala, and G.~Siemens, ``Towards automated content analysis of discussion transcripts: A cognitive presence case,'' in \emph{Proceedings of the {{Sixth International Conference}} on {{Learning Analytics}} \& {{Knowledge}}}, ser. {{LAK}} '16.\hskip 1em plus 0.5em minus 0.4em\relax New York, NY, USA: Association for Computing Machinery, 2016, pp. 15--24.

\bibitem{barbosaAutomaticCrosslanguageClassification2020}
G.~Barbosa, R.~Camelo, A.~P. Cavalcanti, P.~Miranda, R.~F. Mello, V.~Kovanovi{\'c}, and D.~Ga{\v s}evi{\'c}, ``Towards automatic cross-language classification of cognitive presence in online discussions,'' in \emph{Proceedings of the {{Tenth International Conference}} on {{Learning Analytics}} \& {{Knowledge}}}, ser. {{LAK}} '20.\hskip 1em plus 0.5em minus 0.4em\relax New York, NY, USA: Association for Computing Machinery, 2020, pp. 605--614.

\bibitem{liuLookingMOOCDiscussion2022}
Z.~Liu, X.~Kong, S.~Liu, Z.~Yang, and C.~Zhang, ``Looking at {{MOOC}} discussion data to uncover the relationship between discussion pacings, learners' cognitive presence and learning achievements,'' \emph{Education and Information Technologies}, vol.~27, no.~6, pp. 8265--8288, 2022.

\bibitem{yanPracticalEthicalChallenges2023a}
L.~Yan, L.~Sha, L.~Zhao, Y.~Li, R.~{Martinez-Maldonado}, G.~Chen, X.~Li, Y.~Jin, and D.~Ga{\v s}evi{\'c}, ``Practical and ethical challenges of large language models in education: {{A}} systematic scoping review,'' \emph{British Journal of Educational Technology}, 2023.

\bibitem{laskarSystematicStudyComprehensive2023}
M.~T.~R. Laskar, M.~S. Bari, M.~Rahman, M.~A.~H. Bhuiyan, S.~Joty, and J.~Huang, ``A {{Systematic Study}} and {{Comprehensive Evaluation}} of {{ChatGPT}} on {{Benchmark Datasets}},'' in \emph{Findings of the {{Association}} for {{Computational Linguistics}}: {{ACL}} 2023}, A.~Rogers, J.~{Boyd-Graber}, and N.~Okazaki, Eds.\hskip 1em plus 0.5em minus 0.4em\relax Toronto, Canada: Association for Computational Linguistics, 2023, pp. 431--469.

\bibitem{touvronLlamaOpenEfficient2023a}
H.~Touvron, T.~Lavril, G.~Izacard, X.~Martinet, M.-A. Lachaux, T.~Lacroix, B.~Rozi{\`e}re, N.~Goyal, E.~Hambro, F.~Azhar, A.~Rodriguez, A.~Joulin, E.~Grave, and G.~Lample, ``{{LLaMA}}: {{Open}} and {{Efficient Foundation Language Models}},'' 2023.

\bibitem{jiangMistral7B2023}
A.~Q. Jiang, A.~Sablayrolles, A.~Mensch, C.~Bamford, D.~S. Chaplot, D.~de~las Casas, F.~Bressand, G.~Lengyel, G.~Lample, L.~Saulnier, L.~R. Lavaud, M.-A. Lachaux, P.~Stock, T.~L. Scao, T.~Lavril, T.~Wang, T.~Lacroix, and W.~E. Sayed, ``Mistral {{7B}},'' 2023.

\bibitem{ouyangTrainingLanguageModels2022a}
\BIBentryALTinterwordspacing
L.~Ouyang, J.~Wu, X.~Jiang, D.~Almeida, C.~Wainwright, P.~Mishkin, C.~Zhang, S.~Agarwal, K.~Slama, A.~Ray, J.~Schulman, J.~Hilton, F.~Kelton, L.~Miller, M.~Simens, A.~Askell, P.~Welinder, P.~F. Christiano, J.~Leike, and R.~Lowe, ``Training language models to follow instructions with human feedback,'' \emph{Advances in Neural Information Processing Systems}, vol.~35, pp. 27\,730--27\,744, 2022. [Online]. Available: \url{https://proceedings.neurips.cc/paper_files/paper/2022/hash/b1efde53be364a73914f58805a001731-Abstract-Conference.html}
\BIBentrySTDinterwordspacing

\bibitem{sahoo2024systematic}
P.~Sahoo, A.~K. Singh, S.~Saha, V.~Jain, S.~Mondal, and A.~Chadha, ``A systematic survey of prompt engineering in large language models: Techniques and applications,'' \emph{arXiv preprint arXiv:2402.07927}, 2024.

\bibitem{wangCanChatGPTDetect2023a}
D.~Wang, D.~Shan, Y.~Zheng, K.~Guo, G.~Chen, and Y.~Lu, ``Can {{ChatGPT Detect Student Talk Moves}} in {{Classroom Discourse}}? {{A Preliminary Comparison}} with {{Bert}},'' in \emph{Proceedings of the 16th {{International Conference}} on {{Educational Data Mining}}}, 2023, pp. 515--519.

\bibitem{gilardiChatGPTOutperformsCrowd2023a}
F.~Gilardi, M.~Alizadeh, and M.~Kubli, ``{{ChatGPT}} outperforms crowd workers for text-annotation tasks,'' \emph{Proceedings of the National Academy of Sciences of the United States of America}, vol. 120, no.~30, 2023.

\bibitem{sunTextClassificationLarge2023}
X.~Sun, X.~Li, J.~Li, F.~Wu, S.~Guo, T.~Zhang, and G.~Wang, ``Text {{Classification}} via {{Large Language Models}},'' in \emph{Findings of the {{Association}} for {{Computational Linguistics}}: {{EMNLP}} 2023}, H.~Bouamor, J.~Pino, and K.~Bali, Eds.\hskip 1em plus 0.5em minus 0.4em\relax Singapore: Association for Computational Linguistics, 2023, pp. 8990--9005.

\bibitem{gretzZeroshotTopicalText2023a}
S.~Gretz, A.~Halfon, I.~Shnayderman, O.~{Toledo-Ronen}, A.~Spector, L.~Dankin, Y.~Katsis, O.~Arviv, Y.~Katz, N.~Slonim, and L.~{Ein-Dor}, ``Zero-shot {{Topical Text Classification}} with {{LLMs}} - an {{Experimental Study}},'' in \emph{Findings of the {{Association}} for {{Computational Linguistics}}: {{EMNLP}} 2023}, H.~Bouamor, J.~Pino, and K.~Bali, Eds.\hskip 1em plus 0.5em minus 0.4em\relax Singapore: Association for Computational Linguistics, 2023, pp. 9647--9676.

\bibitem{lewisRetrievalAugmentedGenerationKnowledgeIntensive2020a}
\BIBentryALTinterwordspacing
P.~Lewis, E.~Perez, A.~Piktus, F.~Petroni, V.~Karpukhin, N.~Goyal, H.~K{\"u}ttler, M.~Lewis, W.-t. Yih, T.~Rockt{\"a}schel, S.~Riedel, and D.~Kiela, ``Retrieval-{{Augmented Generation}} for {{Knowledge-Intensive NLP Tasks}},'' in \emph{Advances in {{Neural Information Processing Systems}}}, vol.~33.\hskip 1em plus 0.5em minus 0.4em\relax Curran Associates, Inc., 2020, pp. 9459--9474. [Online]. Available: \url{https://proceedings.neurips.cc/paper/2020/hash/6b493230205f780e1bc26945df7481e5-Abstract.html}
\BIBentrySTDinterwordspacing

\bibitem{liu-etal-2022-generated}
\BIBentryALTinterwordspacing
J.~Liu, A.~Liu, X.~Lu, S.~Welleck, P.~West, R.~Le~Bras, Y.~Choi, and H.~Hajishirzi, ``Generated knowledge prompting for commonsense reasoning,'' in \emph{Proceedings of the 60th Annual Meeting of the Association for Computational Linguistics (Volume 1: Long Papers)}, S.~Muresan, P.~Nakov, and A.~Villavicencio, Eds.\hskip 1em plus 0.5em minus 0.4em\relax Dublin, Ireland: Association for Computational Linguistics, May 2022, pp. 3154--3169. [Online]. Available: \url{https://aclanthology.org/2022.acl-long.225}
\BIBentrySTDinterwordspacing

\bibitem{xuEffectsTeacherRole2020}
B.~Xu, N.-S. Chen, and G.~Chen, ``Effects of teacher role on student engagement in {{WeChat-Based}} online discussion learning,'' \emph{Computers \& Education}, vol. 157, p. 103956, 2020.

\bibitem{demszkyGoEmotionsDatasetFineGrained2020}
D.~Demszky, D.~{Movshovitz-Attias}, J.~Ko, A.~Cowen, G.~Nemade, and S.~Ravi, ``{{GoEmotions}}: {{A Dataset}} of {{Fine-Grained Emotions}},'' in \emph{Proceedings of the 58th {{Annual Meeting}} of the {{Association}} for {{Computational Linguistics}}}, D.~Jurafsky, J.~Chai, N.~Schluter, and J.~Tetreault, Eds.\hskip 1em plus 0.5em minus 0.4em\relax {Online}: {Association for Computational Linguistics}, 2020, pp. 4040--4054.

\bibitem{blacheTwolevelClassificationDialogue2020}
P.~Blache, M.~Abderrahmane, S.~Rauzy, M.~Ochs, and H.~Oufaida, ``Two-level classification for dialogue act recognition in task-oriented dialogues,'' in \emph{Proceedings of the 28th {{International Conference}} on {{Computational Linguistics}}}, D.~Scott, N.~Bel, and C.~Zong, Eds.\hskip 1em plus 0.5em minus 0.4em\relax Barcelona, Spain (Online): International Committee on Computational Linguistics, 2020, pp. 4915--4925.

\bibitem{almatrafiNeedleHaystackIdentifying2018a}
O.~Almatrafi, A.~Johri, and H.~Rangwala, ``Needle in a haystack: {{Identifying}} learner posts that require urgent response in {{MOOC}} discussion forums,'' \emph{Computers \& Education}, vol. 118, pp. 1--9, 2018.

\bibitem{sha2022leveraging}
L.~Sha, M.~Rakovi{\'c}, A.~Das, D.~Ga{\v{s}}evi{\'c}, and G.~Chen, ``Leveraging class balancing techniques to alleviate algorithmic bias for predictive tasks in education,'' \emph{IEEE Transactions on Learning Technologies}, vol.~15, no.~4, pp. 481--492, 2022.

\bibitem{hoyEducationalPsychology2016}
A.~W. Hoy, \emph{Educational Psychology}, 13th~ed., ser. Always Learning.\hskip 1em plus 0.5em minus 0.4em\relax {Boston Columbus Indianapolis New York San Francisco}: {Pearson}, 2016.

\bibitem{liuSentimentAnalysisMining2020}
B.~Liu, \emph{Sentiment {{Analysis}}: {{Mining Opinions}}, {{Sentiments}}, and {{Emotions}}}.\hskip 1em plus 0.5em minus 0.4em\relax {Cambridge University Press}, 2020-10-15.

\bibitem{leeFewshotEnoughExploring2023}
U.~Lee, H.~Jung, Y.~Jeon, Y.~Sohn, W.~Hwang, J.~Moon, and H.~Kim, ``Few-shot is enough: Exploring {{ChatGPT}} prompt engineering method for automatic question generation in english education,'' \emph{Education and Information Technologies}, 2023.

\bibitem{chevrierExploringAntecedentsConsequences2019}
M.~Chevrier, K.~R. Muis, G.~J. Trevors, R.~Pekrun, and G.~M. Sinatra, ``Exploring the antecedents and consequences of epistemic emotions,'' \emph{Learning and Instruction}, vol.~63, p. 101209, 2019.

\bibitem{chiICAPFrameworkLinking2014}
M.~T.~H. Chi and R.~Wylie, ``The {{ICAP Framework}}: {{Linking Cognitive Engagement}} to {{Active Learning Outcomes}},'' \emph{Educational Psychologist}, vol.~49, no.~4, pp. 219--243, 2014.

\bibitem{atapattuDetectingCognitiveEngagement2019}
T.~Atapattu, M.~Thilakaratne, R.~Vivian, and K.~Falkner, ``Detecting cognitive engagement using word embeddings within an online teacher professional development community,'' \emph{Computers \& Education}, vol. 140, p. 103594, 2019.

\bibitem{garrisonCriticalThinkingCognitive2001b}
D.~R. Garrison, T.~Anderson, and W.~Archer, ``Critical thinking, cognitive presence, and computer conferencing in distance education,'' \emph{American Journal of Distance Education}, vol.~15, no.~1, pp. 7--23, 2001.

\bibitem{netoAutomaticContentAnalysis2021}
V.~Neto, V.~Rolim, A.~Pinheiro, R.~D. Lins, D.~Gasevic, and R.~F. Mello, ``Automatic {{Content Analysis}} of {{Online Discussions}} for {{Cognitive Presence}}: {{A Study}} of the {{Generalizability Across Educational Contexts}},'' \emph{IEEE Transactions on Learning Technologies}, vol.~14, no.~3, pp. 299--312, 2021.

\bibitem{brown2020language}
T.~Brown, B.~Mann, N.~Ryder, M.~Subbiah, J.~D. Kaplan, P.~Dhariwal, A.~Neelakantan, P.~Shyam, G.~Sastry, A.~Askell \emph{et~al.}, ``Language models are few-shot learners,'' \emph{Advances in neural information processing systems}, vol.~33, pp. 1877--1901, 2020.

\bibitem{chawla2002smote}
N.~V. Chawla, K.~W. Bowyer, L.~O. Hall, and W.~P. Kegelmeyer, ``Smote: synthetic minority over-sampling technique,'' \emph{Journal of artificial intelligence research}, vol.~16, pp. 321--357, 2002.

\bibitem{zheng2023judging}
L.~Zheng, W.-L. Chiang, Y.~Sheng, S.~Zhuang, Z.~Wu, Y.~Zhuang, Z.~Lin, Z.~Li, D.~Li, E.~P. Xing, H.~Zhang, J.~E. Gonzalez, and I.~Stoica, ``Judging llm-as-a-judge with mt-bench and chatbot arena,'' 2023.

\bibitem{liuMOOCBERTAutomaticallyIdentifying2023}
Z.~Liu, X.~Kong, H.~Chen, S.~Liu, and Z.~Yang, ``{{MOOC-BERT}}: {{Automatically Identifying Learner Cognitive Presence From MOOC Discussion Data}},'' \emph{IEEE Transactions on Learning Technologies}, vol.~16, no.~4, pp. 528--542, 2023.

\bibitem{metaIntroducingMetaLlama}
\BIBentryALTinterwordspacing
Meta, ``Introducing {{Meta Llama}} 3: {{The}} most capable openly available {{LLM}} to date.'' [Online]. Available: \url{https://ai.meta.com/blog/meta-llama-3/}
\BIBentrySTDinterwordspacing

\bibitem{jiang2024mixtral}
A.~Q. Jiang, A.~Sablayrolles, A.~Roux, A.~Mensch, B.~Savary, C.~Bamford, D.~S. Chaplot, D.~de~las Casas, E.~B. Hanna, F.~Bressand, G.~Lengyel, G.~Bour, G.~Lample, L.~R. Lavaud, L.~Saulnier, M.-A. Lachaux, P.~Stock, S.~Subramanian, S.~Yang, S.~Antoniak, T.~L. Scao, T.~Gervet, T.~Lavril, T.~Wang, T.~Lacroix, and W.~E. Sayed, ``Mixtral of experts,'' 2024.

\bibitem{devlinBERTPretrainingDeep2019}
J.~Devlin, M.-W. Chang, K.~Lee, and K.~Toutanova, ``{{BERT}}: {{Pre-training}} of {{Deep Bidirectional Transformers}} for {{Language Understanding}},'' in \emph{Proceedings of the 2019 {{Conference}} of the {{North American Chapter}} of the {{Association}} for {{Computational Linguistics}}: {{Human Language Technologies}}, {{Volume}} 1 ({{Long}} and {{Short Papers}})}.\hskip 1em plus 0.5em minus 0.4em\relax Minneapolis, Minnesota: Association for Computational Linguistics, 2019, pp. 4171--4186.

\bibitem{liu2019roberta}
Y.~Liu, M.~Ott, N.~Goyal, J.~Du, M.~Joshi, D.~Chen, O.~Levy, M.~Lewis, L.~Zettlemoyer, and V.~Stoyanov, ``Roberta: A robustly optimized bert pretraining approach,'' \emph{arXiv preprint arXiv:1907.11692}, 2019.

\bibitem{karlTransformersAreShortText2023}
F.~Karl and A.~Scherp, ``Transformers are {{Short-Text Classifiers}},'' in \emph{Machine {{Learning}} and {{Knowledge Extraction}}}, ser. Lecture {{Notes}} in {{Computer Science}}, A.~Holzinger, P.~Kieseberg, F.~Cabitza, A.~Campagner, A.~M. Tjoa, and E.~Weippl, Eds.\hskip 1em plus 0.5em minus 0.4em\relax Cham: Springer Nature Switzerland, 2023, pp. 103--122.

\bibitem{zhaoChatAgriExploringPotentials2023a}
B.~Zhao, W.~Jin, J.~Del~Ser, and G.~Yang, ``{{ChatAgri}}: {{Exploring}} potentials of {{ChatGPT}} on cross-linguistic agricultural text classification,'' \emph{Neurocomputing}, vol. 557, p. 126708, 2023.

\end{thebibliography}
\bibliographystyle{IEEEtran}

\begin{IEEEbiography}
[{\includegraphics[width=1in,height=1.25in,clip,keepaspectratio]{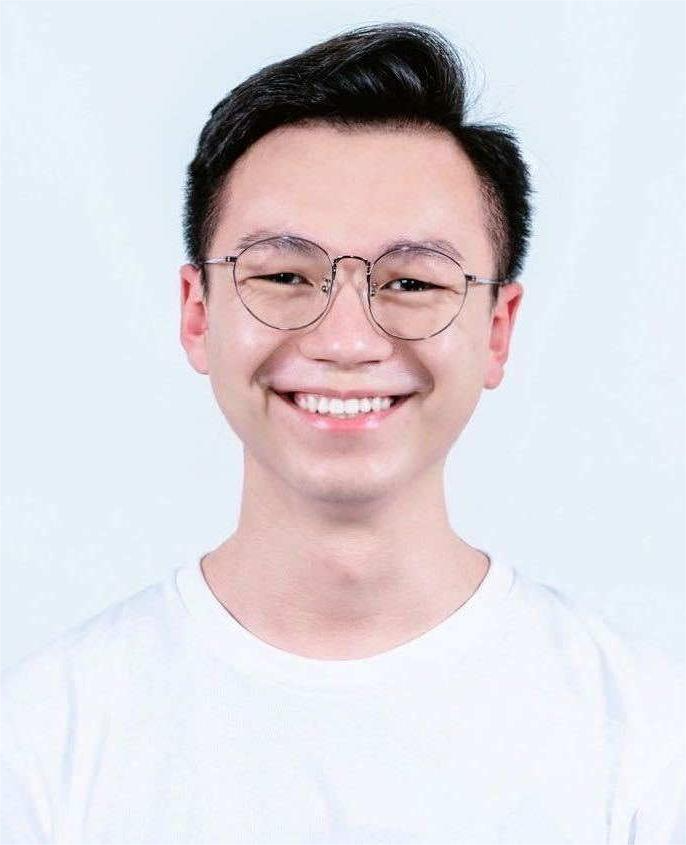}}]{Shiqi Liu}
received M.S. degree in education information science and technology from Central China Normal University (CCNU), Wuhan, China, in 2021.

He is currently working toward the Ph.D. degree in educational technology from the Faculty of Artificial Intelligence in Education, CCNU. He is a visiting PhD student at the Centre for Learning Analytics, Faculty of Information Technology, Monash University. 

His research interests include natural language processing and learning analytics. His work has been published in the prestigious international AI and educational technology journals and conferences \textit{Computers \& Education}, \textit{Knowledge-Based Systems}, and \textit{IJCAI 2024}.

\end{IEEEbiography}

\begin{IEEEbiography}
[{\includegraphics[width=1in,height=1.25in,clip,keepaspectratio]{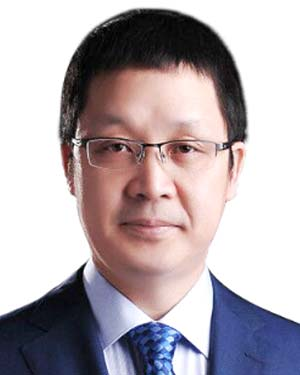}}]{Sannyuya Liu}
(Member, IEEE) received the Ph.D. degree in systems engineering from the Huazhong University of Science and Technology, Wuhan, China, in 2003. 

He is currently a Professor and a Ph.D. Supervisor with the National Engineering Research Center of Educational Big Data, CCNU, Wuhan. 

His research interests include computer application, artificial intelligence, and educational information technology. He is a member of the Institute of Electrical and Electronics Engineers. 
\end{IEEEbiography}

\begin{IEEEbiography}
[{\includegraphics[width=1in,height=1.25in,clip,keepaspectratio]{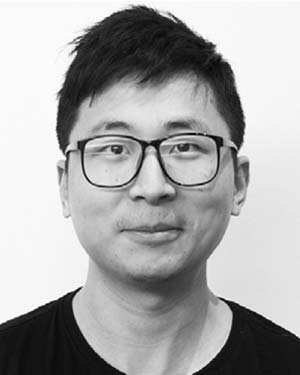}}]{Lele Sha}
 is currently a Research fellow in the Centre for Learning Analytics at Monash (CoLAM), an interdisciplinary group of Education and information technology (i.e., Educational Technology). 

Lele's research is centered on ethical and practical implications of natural language processing technology in education. Lele has published numerous peer-reviewed articles in respected venues related to natural language processing, responsible AI, and educational technology.

\end{IEEEbiography}

\begin{IEEEbiography}
[{\includegraphics[width=1in,height=1.25in,clip,keepaspectratio]{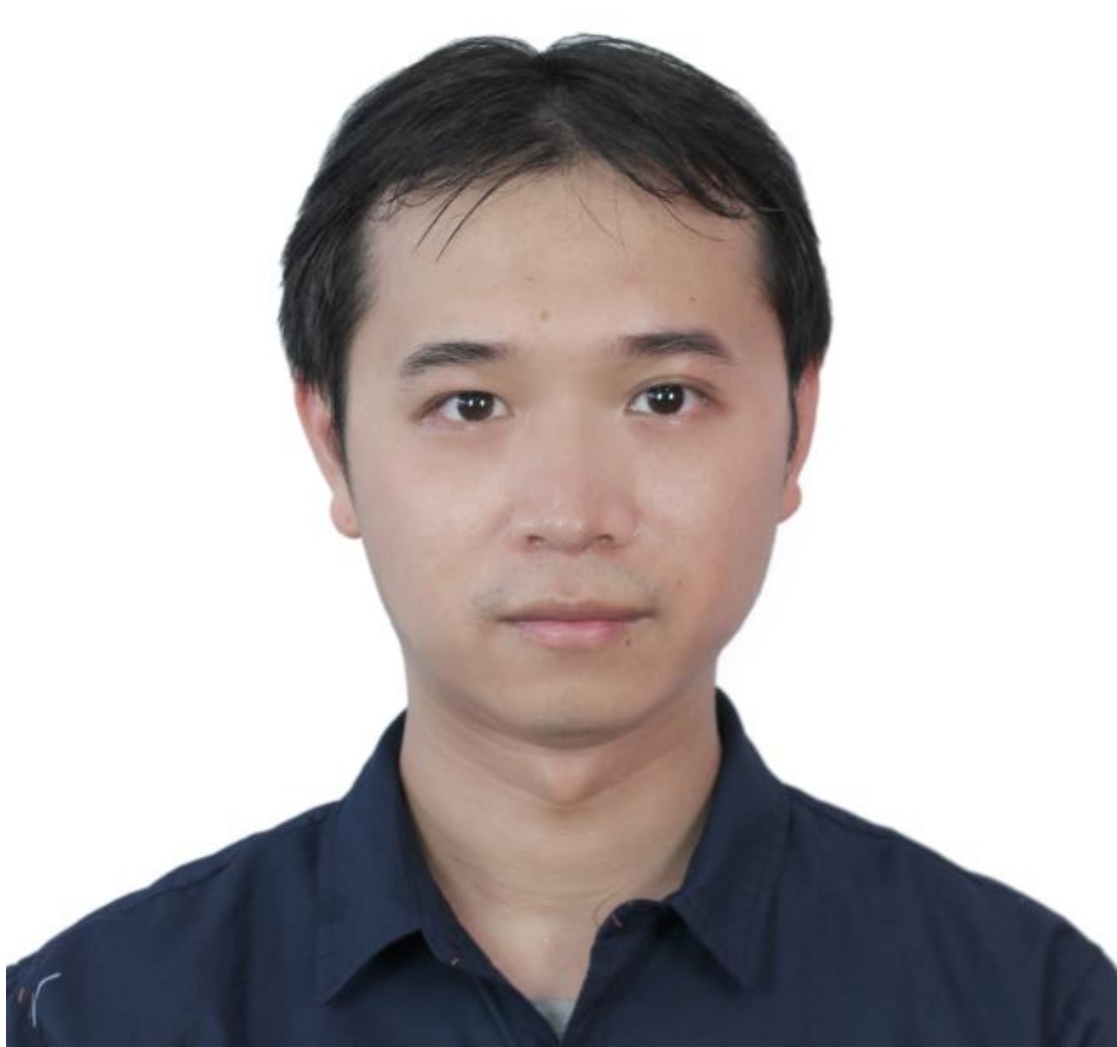}}]{Zijie Zeng} is currently pursuing his PhD at Monash University in Australia and is expected to receive his degree by 2025. His research and work focus on natural language processing, recommendation techniques, and data mining. Recently, he has been working on a research project about AI-generated text detection, and his findings were published at the prestigious international AI conferences AAAI 2024 and IJCAI 2024.
\end{IEEEbiography}

\begin{IEEEbiography}
[{\includegraphics[width=1in,height=1.25in,clip,keepaspectratio]{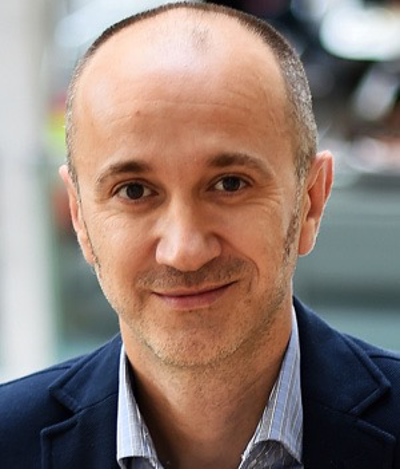}}]{Dragan Ga\v{s}evi\'{c}}
received the Ph.D. degree in computer science from the University of Belgrade, Beograd, Serbia, in 2005. 

He is currently a Distinguished Professor of Learning Analytics with the Faculty of Information Technology and the Director of the Centre for Learning Analytics, Monash University, Clayton, VIC, Australia. He is a frequent keynote speaker and a (co-) author of numerous research papers and books. His research interests include self-regulated and social learning, higher education policy, and data mining. 

Prof. Ga\v{s}evi\'{c} was the past President from 2015 to 2017 and a cofounder of the Society for Learning Analytics Research, a founding Program Chair of the International Conference on Learning Analytics and Knowledge, and a Founding Editor of the Journal of Learning Analytics.
\end{IEEEbiography}

\begin{IEEEbiography}
[{\includegraphics[width=1in,height=1.25in,clip,keepaspectratio]{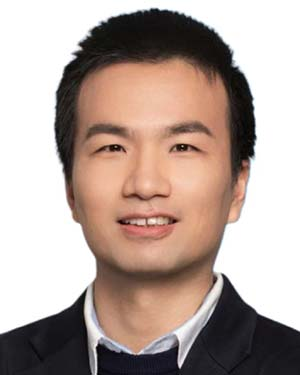}}]{Zhi Liu}
received the Ph.D. degree in education information science and technology from CCNU, Wuhan, China, in 2014. 

He is an associate researcher in National Engineering Research Center of Educational Big Data at CCNU. His focuses include learning analytics and educational data mining. He was a guest researcher in the Department of Computer Science at Humboldt University of Berlin from 2017-2018. 

He is currently organizing committee chair of ICET. As a PI, he is leading the multi research projects of NSFC about Artificial Intelligence in Education. He won the first Prize of the Teaching Achievement Award of Higher Education Institutions in Hubei Province in 2022, the Special Prize of the Science and Technology Progress Award of the Chinese Association of Automation in 2020.
\end{IEEEbiography}

\vfill

\end{document}